%% file: main.tex
\documentclass[10pt,twocolumn,letterpaper]{article}

\usepackage[pagenumbers]{cvpr} %

\usepackage[table]{xcolor}

\input{preamble}
\definecolor{cvprblue}{rgb}{0.21,0.49,0.74}
\usepackage[pagebackref,breaklinks,colorlinks,citecolor=cvprblue]{hyperref}

\usepackage{textcomp, gensymb}
\usepackage{paralist}
\usepackage{chngcntr}

\newcommand*{\circled}[1]{\lower.7ex\hbox{\tikz\draw (0pt, 0pt)%
    circle (.5em) node {\makebox[1em][c]{\small #1}};}}
\newcommand{\best}[1]{\textbf{#1}}

\expandafter\def\expandafter\normalsize\expandafter{%
    \normalsize%
    \setlength\abovedisplayskip{4pt}%
    \setlength\belowdisplayskip{4pt}%
    \setlength\abovedisplayshortskip{-8pt}%
    \setlength\belowdisplayshortskip{2pt}%
}

\def\paperID{186} %
\def\confName{3DV\xspace}
\def\confYear{2025\xspace}

\title{Synthesizing Consistent Novel Views via 3D Epipolar Attention \\ without Re-Training}

\author{
Botao Ye$^{1,2}$ \quad Sifei Liu$^3$ \quad Xueting Li$^3$ \quad Marc Pollefeys$^{1,4}$ \quad Ming-Hsuan Yang$^{5}$ \\
$^1$ETH Zurich \quad $^2$ETH AI Center \quad $^3$NVIDIA \quad 
$^4$Microsoft \quad $^5$UC Merced
{}
{}
}

\begin{document}

\twocolumn[{
\renewcommand\twocolumn[1][]{#1}
\maketitle
\vspace{-12mm}
\begin{center}
    \centering
    \captionsetup{type=figure}
    \includegraphics[width=.8\textwidth]{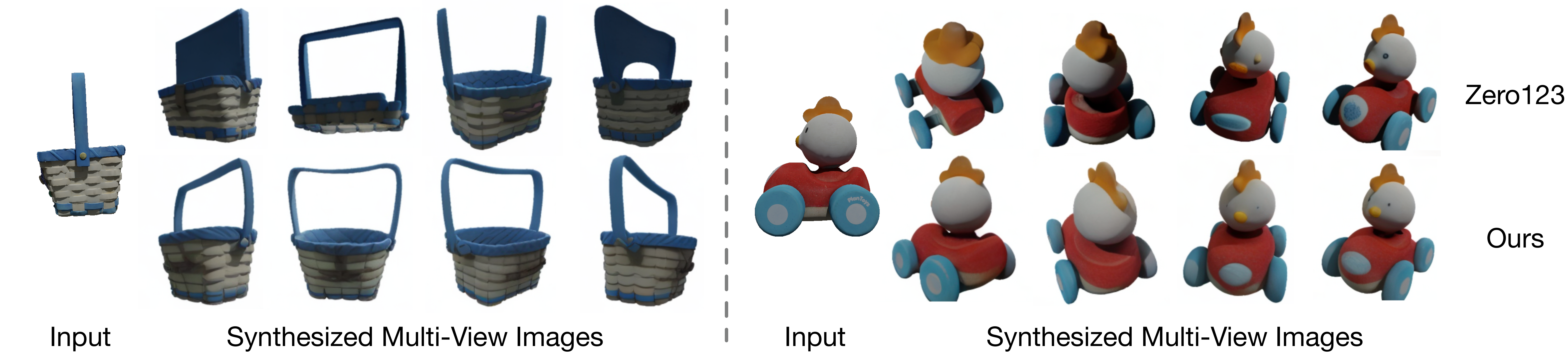}
    \vspace{-2mm}
    \captionof{figure}{Given an input image and a sequence of relative camera pose transformations, our method synthesizes more consistent novel view images. Our method \textit{does not need to re-train} the baseline model (Zero123) and \textit{supports arbitrary relative camera poses}.}
    \vspace{-1mm}
\end{center}
}]

\maketitle

\input{sec/0_abstract}    
\input{sec/1_intro}
\input{sec/2_related}
\input{sec/3_method}

\input{sec/4_experiments}

\input{sec/5_conclusion}

{
    \small
    \bibliographystyle{ieeenat_fullname}
    \bibliography{main}
}
\input{sec/X_suppl}

\end{document}

%% file: sec/0_abstract.tex
\begin{abstract}
\vspace{-2mm}
Large diffusion models demonstrate remarkable zero-shot capabilities in novel view synthesis from a single image. 
However, these models often face challenges in maintaining consistency across novel and reference views.
A crucial factor leading to this issue is the limited utilization of contextual information from reference views.
Specifically, when there is an overlap in the viewing frustum between two views, it is essential to ensure that the corresponding regions maintain consistency in both geometry and appearance.
This observation leads to a simple yet effective approach, where we propose to use epipolar geometry to locate and retrieve overlapping information from the input view. This information is then incorporated into the generation of target views, eliminating the need for training or fine-tuning, as the process requires no learnable parameters.
Furthermore, to enhance the overall consistency of generated views, we extend the utilization of epipolar attention to a multi-view setting, allowing retrieval of overlapping information from the input view and other target views.
Qualitative and quantitative experimental results demonstrate the effectiveness of our method in significantly improving the consistency of synthesized views without the need for any fine-tuning.
Moreover, This enhancement also boosts the performance of downstream applications such as 3D reconstruction. The code is available at \url{https://github.com/botaoye/ConsisSyn}.
\end{abstract}

%% file: sec/1_intro.tex
\vspace{-6mm}

\section{Introduction}
\label{sec:intro}

Synthesizing high-quality novel view images from a single input image is a long-standing and challenging problem. It requires inheriting the appearance of objects in the observed regions of the input image while also hallucinating unseen regions. Recent studies~\cite{3dim, zero123} approach this problem as an image-to-image translation task and implement it using diffusion models~\cite{diffusion, ddim}, drawing inspiration from their successful application in 2D image generation~\cite{stable-diffusion, imagen}.	
While they exhibit remarkable zero-shot capabilities when trained with large-scale 2D and 3D datasets, they still face challenges in maintaining 3D consistency between the target view and the generated multi-view images, due to the probabilistic nature of diffusion models. This limitation adversely affects downstream applications such as 3D reconstruction~\cite{dreamfusion, neus}.

In this paper, we propose to improve the consistency of synthesized multi-view images by optimizing the utilization of reference image information. 
Notably, maintaining consistency between the generated image and the corresponding observed regions in the input view is a crucial requirement in the task of single-image conditioned novel view synthesis. 
However, existing methods often overlook this constraint by merely considering the input image as a condition or network input, which fails to guarantee such consistency. 
One straightforward method to fulfill this constraint is by warping the content from the input to the target view and subsequently conducting outpainting for the remaining regions~\cite{text2nerf, xiang20233d}.
However, 3D warping relies on precise depth information, which is hard to obtain.
Additionally, direct warping struggles with occlusion and illumination variations across different views.

\begin{figure}[t]
\vspace{-5mm}
    \centering
    \includegraphics[width=.7\linewidth]{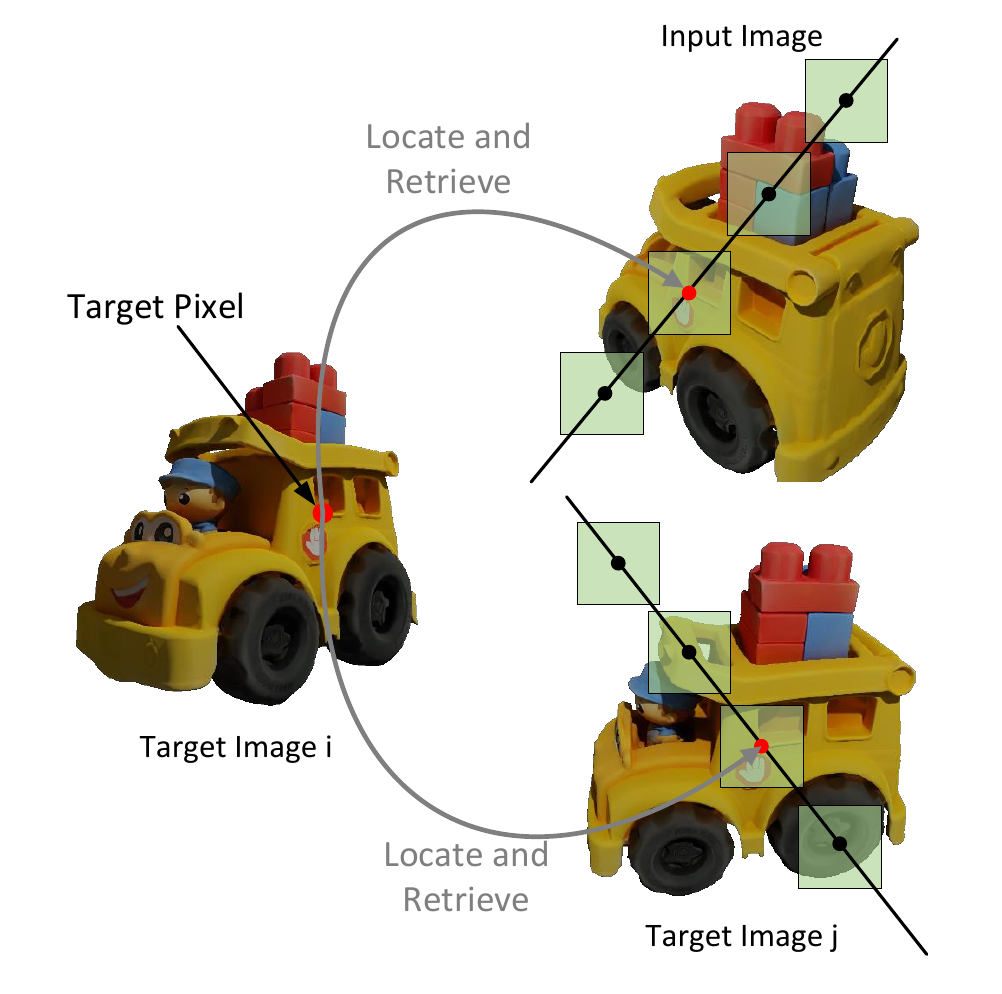}
     \vspace{-5mm}
    \caption{When the camera viewing frustum of two views overlaps, for a point on one of the images, we can find its correspondence on the epipolar line of the other view.}
    \label{fig:epipolar_vis}
    \vspace{-3mm}
\end{figure}

We aim to utilize this constraint to improve the consistency in a more adaptable way.
Despite the intricacies of obtaining depth, we can still reduce the search space for locating corresponding points by incorporating other 3D geometric priors. 
As depicted in Fig.~\ref{fig:epipolar_vis}, the corresponding points in the reference views must be on the epipolar line.
Therefore, we propose an epipolar attention module to locate and gather contextual information. 
For each point in the target view visible in the reference view, we can first constrain the corresponding point to its respective epipolar line in the reference image. Subsequently, we ascertain the corresponding location along the epipolar line by feature matching.
The features at the localized positions are then retrieved and used to constrain the target view generation.

More specifically, we first perform DDIM inversion on the input view and reconstruct the input image using the initial noise provided by the DDIM inversion.
This process yields intermediate features of the input view, which can then be employed to constrain the generation of target views.
Then, in the epipolar attention module, we traverse the corresponding epipolar line in the input view for every point within the target view. During this process, we compute the similarity between the features of the target point and those sampled from the input view. This similarity score is then used to aggregate the corresponding features from the input view.
This soft operation is more adept at handling complex scenarios, such as occlusion (detailed analysis can be found in the Supplementary Material).
Additionally, to avoid any parameter training or fine-tuning, we employ a simple parameter duplication strategy, \ie, we copy all parameters directly from the self-attention layer to obtain the epipolar attention parameters.
To further improve the consistency between different target views, we expand the application of epipolar attention to a multi-view context.
Specifically, we generate multiple target views in an auto-regressive manner. 
When generating a specific novel view, we consider the input view and previously generated target views close to the current viewpoint as context views.
We employ epipolar attention to aggregate overlapping information from all context views, rather than solely from the input view, thereby improving consistency among all generated views.
It is worth mentioning that our epipolar attention reduces the search space compared to locating corresponding points in the full image. Therefore, it requires much less memory when retrieving information from multiple views, making it more friendly to GPUs with small memory capacity.

\input{tables/intro}

We conduct experiences on the Google Scanned Objects~\cite{GSO} dataset to verify the zero-shot novel view synthesis capability and evaluate our method on both generated image quality and the view consistency~\cite{3dim}. 
Additionally, we apply our method to the downstream 3D reconstruction task~\cite{neus} and compare it against the mesh constructed by our baseline model.

The main contributions of this work are:
\begin{compactitem}
    \item  We propose a novel epipolar attention method to locate and retrieve the corresponding information in the reference view, which is then inserted into the generation process of the target view to enhance the consistency between multi-view images.
    \item Experimental results show that our method effectively improves the consistency of the synthesized multi-view images without any training or fine-tuning while maintaining the quality of the generated images.
    \item We apply the synthesized multi-view images to a downstream 3D reconstruction task, and the results show that the more consistent images further improve the 3D reconstruction results.
\end{compactitem} 

%% file: tables/intro.tex
\newcommand{\rot}[2][l]{\rotatebox[origin=#1]{90}{#2}} 
\definecolor{checkyes}{rgb}{0.7, 1.0, 0.7}
\definecolor{crossno}{rgb}{1.0, 0.7, 0.7}
\def \y {$\checkmark$\cellcolor{checkyes}}
\def \n {$\times$\cellcolor{crossno}}

\begin{table}
\centering
\setlength{\tabcolsep}{3pt}
\def\arraystretch{1}
\resizebox{0.8\linewidth}{!}{
\begin{tabular}{lc|ccc|ccc|c}
\toprule
 & Zero123
 & \multicolumn{3}{c}{Fixed Camera}
 & \multicolumn{3}{c}{Additional Net}
 & Ours
 \\
 \cmidrule(lr){2-2}
 \cmidrule(lr){3-5}
 \cmidrule(lr){6-8}
 \cmidrule(lr){9-9}
 &
  \rot{Zero123~\cite{zero123}} &
  
  \rot{SyncDreamer~\cite{SyncDreamer}} &
  \rot{MVDream~\cite{mvdream}} &  
  \rot{Zero123++~\cite{zero123++}} &  

  \rot{SparseFusion~\cite{sparsefusion}} &
  \rot{PGD~\cite{tseng2023consistent}} &
  \rot{Consistent123~\cite{ye2023consistent}} &

    \rot{Ours}\\
\cmidrule{2-9}
{1) S.V. Condition}       & \y & \y & \y & \y & \n & \y & \y & \y \\
{2) Generalizability}     & \y & \y & \y & \y & \n & \n & \y & \y \\
{3) No Extra Training}   & \n & \n & \n & \n & \n & \n & \n & \y \\
{4) Multi-view Consis}    & \n & \y & \y & \y & \y & \y & \y & \y \\
{5) Free Trajectory}      & \y & \n & \n & \n & \y & \y & \y & \y \\
\bottomrule
\end{tabular}
}
\vspace{-.5em}
\captionof{table}{
    \textbf{Comparison with related methods.}
    Each row represents the ability to:
    2) generalize well to arbitrary objects,
    3) work without requiring extra retraining,
    4) generate multi-view consistent images,
    and 5) generate images in arbitrary camera poses.
    See Sec.~\ref{sec:attn_analysis} for a detailed comparison.
}
\label{tab:intro}
\vspace{-5mm}
\end{table}

%% file: sec/2_related.tex
\section{Related Work}
\label{sec:related_work}

\textbf{Diffusion Models for Novel View Synthesis.}
Diffusion models show impressive results on the text-to-image task~\cite{stable-diffusion, imagen, dalle2}. 
Therefore, a line of work aims to extend it to the novel view synthesis (NVS) task, where they generate novel view images based on reference images and desired relative camera poses.
Such synthesized multi-view images find utility in various applications such as distillation purposes~\cite{dreamfusion, sjc, magic3d, fantasia3d, prolificdreamer}, or for directly training NeRF-like 3D assets~\cite{nerf, neus, sparseneus}.
3DiM~\cite{3dim} implements this idea by training a diffusion model conditioned on reference images and relative camera poses.
SparseFusion~\cite{sparsefusion} and GeNVS~\cite{genvs} first generate course latent feature of the target view as additional input to the diffusion model.
However, these methods are trained on objects from specific classes or relatively small datasets, making it challenging to generalize to arbitrary objects.
Zero123~\cite{zero123} obtains impressive zero-shot generalizability by fine-tuning a 2D diffusion model, \ie, Stable Diffusion~\cite{stable-diffusion}, on a large-scale 3D rendered dataset~\cite{objaverse}.
However, novel view images generated by Zero123 can suffer from consistency problems, especially when relatively large pose transformations are present.	
To address this issue, some very recent studies~\cite{SyncDreamer, mvdream, zero123++, consistent123, ye2023consistent} try to add additional modules and fine-tune the Zero123 or LDM model to obtain better consistency, which requires significant computational resources.
In contrast to these approaches, we focus on enhancing the consistency of pre-trained models without the need for any fine-tuning. Tab.~\ref{tab:intro} provides an overview comparison, while Sec.~\ref{sec:attn_analysis} offers a detailed comparison.

\noindent\textbf{Image-to-Image Translation.}
Image-to-image translation (I2I) involves learning a mapping from an input image to an output image while preserving specific properties like the scene layout or object structure. 
Our paper's primary focus can be viewed as an I2I task, where the condition is the pose, aiming to transform the input image to the desired pose.
One of the main challenges in the pose-guided novel view generation task is maintaining consistency between the target images and the input image. 
This challenge shares similarities with the issues encountered in text-guided image-to-image translation tasks~\cite{p2p, pnp, masactrl, imagic, dragondiffusion}.
For instance, works such as~\cite{p2p, pnp, masactrl} manipulate self-attention, cross-attention, or spatial features within the U-Net~\cite{unet} structure to preserve the desired concept in the input image.
However, these methods primarily target 2D image translation or editing tasks, lacking 3D structural information and struggling to discern what to preserve or discard in the context of the NVS task. 
In contrast, our method incorporates 3D geometry information into the translation process to better preserve the desired information in the input view.

\noindent\textbf{Epipolar Geometry in DNN.}
Epipolar geometry is used in many previous works~\cite{tseng2023consistent, mvster, he2020epipolar, ecsic, suhail2022generalizable}.
They often integrate epipolar geometry into network modules and employ it for network training.
In contrast, we use the epipolar geometry to generate images without training or fine-tuning to localize better and retrieve the corresponding information using the features from a trained diffusion model.

%% file: sec/3_method.tex
\section{Preliminaries}
\label{sec:background}

\begin{figure*}[t]
    \vspace{-3mm}
    \centering
    \includegraphics[width=0.8\linewidth]{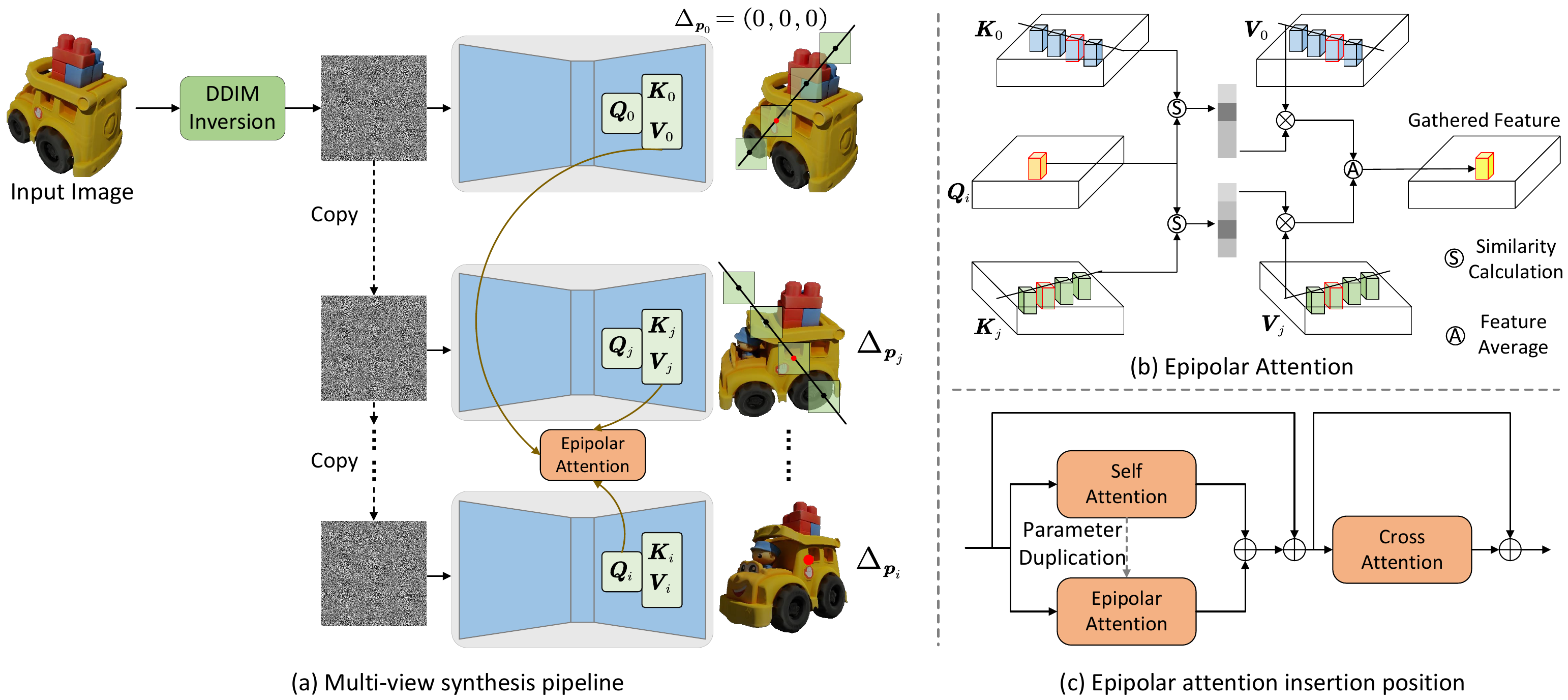}
    \vspace{-2mm}
    \caption{\textbf{Overview of our method.} 
    (a) We first perform DDIM inversion on the input image to obtain the initial noise, which is shared during the multi-view image generation process. Throughout the generation of each view, our epipolar attention block efficiently locates and retrieves corresponding information from both the input image and other target views.
    (b) The architecture of our 3D epipolar attention module.
    (c) Location of our inserted epipolar attention block.}
    \label{fig:arch}
    \vspace{-4mm}
\end{figure*}

In this section, we revisit the pose-conditional diffusion model used in our approach (Sec.~\ref{sec:pcdm}), and the DDIM inversion technique used to invert the reference image back to the initial Gaussian noise (Sec.~\ref{sec:ddim}).

\subsection{Pose-Conditioned Diffusion Model} \label{sec:pcdm}
Diffusion models~\cite{sohl2015deep, diffusion, ddim, dhariwal2021diffusion} are probabilistic generative models, which transform an initial Gaussian noise $\boldsymbol{x}_T \sim \mathcal{N}(0, \mathbf{I})$ into an arbitrary meaningful data distribution.
During training, the diffusion \textit{forward} process is applied, in which Gaussian noise is added to the clean data $\boldsymbol{x}_0$ (image in our case):
\begin{equation}
\label{eq:diffusion_forward}
    \boldsymbol{x}_t=\sqrt{\alpha_t} \cdot \boldsymbol{x}_0+\sqrt{1-\alpha_t} \cdot \boldsymbol{z},
\end{equation}
where $\boldsymbol{z} \sim \mathcal{N}(0, \mathbf{I})$ is the random noise and $\{\alpha_t\}, t \in [0, T]$ is the noise schedule indexed by time step $t$.
During inference, the \textit{backward} diffusion process is utilized to progressively denoise $\boldsymbol{x}_T$ to obtain the clean data. This denoising process is facilitated by a neural network $\boldsymbol{\epsilon}_\theta\left(\boldsymbol{x}_t, t\right)$, which predicts noise at each step.

We focus on employing the diffusion model for synthesizing novel views from a single input view. 
This can be seen as an image-to-image translation process that transforms the original image into a novel view image based on their relative camera pose transformation.
Formally, given the reference view image $\boldsymbol{x}_r$ and the relative camera pose transformation $\Delta\boldsymbol{p} = (\mathbf{R}, \mathbf{T})$ between the reference view and the target view, the denoising network predicts noise conditioned on both $\boldsymbol{x}_r$ and $\Delta\boldsymbol{p}$, denoted as:
\begin{equation}
\label{eq:denoise}
    \boldsymbol{z}_t = \boldsymbol{\epsilon}_\theta\left(\boldsymbol{x}_t, t \mid \boldsymbol{x}_r, \Delta\boldsymbol{p}\right).
\end{equation}

In this work, we leverage a pre-trained pose-conditioned diffusion model (Zero123~\cite{zero123}), which in turn is fine-turned from a Latent Diffusion Model~\cite{stable-diffusion}. 
The network is implemented using a U-Net~\cite{unet} structure, consisting of several residual blocks~\cite{resnet}, self-attention blocks, and cross-attention blocks~\cite{transformer}.
At each time step $t$, the feature maps from the previous layer $l-1$ are first feeded in the residual block to obtain feature $\boldsymbol{F}^l_t$.
Subsequently, projection layers are employed to generate distinct query $\boldsymbol{Q}$, key $\boldsymbol{K}$, and value $\boldsymbol{V}$ feature maps (for simplicity, excluding time step $t$ and layer index $l$).
The output feature of the self-attention block, denoted as $\hat{\boldsymbol{F}}$, is computed using the operation $\hat{\boldsymbol{F}} = \boldsymbol{A} \cdot \boldsymbol{V}$, where the attention matrix $\boldsymbol{A}$ is determined as follows:
\begin{equation}
\label{eq:attn_mat}
    \boldsymbol{A}=\operatorname{Softmax}\left(\frac{\boldsymbol{Q} \boldsymbol{K}^\top}{\sqrt{d}}\right),
\end{equation}
where $d$ is the feature dimension.

\subsection{DDIM Inversion} \label{sec:ddim}
\label{subsec:ddim_inversion}
During \textit{backward} diffusion, deterministic DDIM sampling~\cite{ddim} is commonly used to convert noise $\boldsymbol{x}_T$ into clean data $\boldsymbol{x}_0$.
In contrast, DDIM inversion~\cite{ddim, dhariwal2021diffusion} converts the original clean image data $\boldsymbol{x}_0$ back to Gaussian noise $\boldsymbol{x}_T$ by incrementally adding the noise predicted by the network $\boldsymbol{\epsilon}_\theta$.
We also employ DDIM inversion to convert the input image to its initial noise $\boldsymbol{x}^R$. Throughout this conversion, we utilize the input image feature and a fixed relative pose transformation of $[0, 0, 0]$ as the network condition.

\section{Approach}
As mentioned above, our goal is to improve the consistency of the synthetic multi-view image by locating and retrieving the corresponding information (features) in the reference view that overlap with the target view and then using the retrieved features to constrain the target view generation process.
Therefore, we first explicate the methodology for computing the epipolar line and sampling points along it to effectively reduce the searching space concerning the corresponding locations (Sec.~\ref{sec:epi_sample}).
Then, we will describe how to locate the corresponding locations along the epipolar line (Sec.~\ref{sec:fine_locate}). The general idea is to find the correspondence between a point in the target view and sampled points on the epipolar line through feature similarity. To better obtain the similarity information, we analyze the attributes of different features in the U-Net block and then find the appropriate features used to compute the similarity.
Next, we introduce the parameter duplication strategy that facilitates the training-free module, and how to inject the retrieved reference features to constrain the generation process of the target views (Sec.~\ref{sec:feature_use}). 
Finally, we will provide a detailed analysis of our epipolar attention (Sec.~\ref{sec:attn_analysis}).
Fig.~\ref{fig:arch} shows the overall framework of our tuning-free multi-view epipolar attention that enables consistent novel view synthesis.

\subsection{Point Sampling from Epipolar Lines}\label{sec:epi_sample}
Ideally, with the target view's depth map, we can accurately find its correspondence in the reference view by un-projecting each point to 3D space and then re-projecting it into the reference view.
However, obtaining an accurate depth value for arbitrary real-world objects remains challenging, if not infeasible. 
Alternatively, when considering a point $\boldsymbol{p}_i$ in the target view, its corresponding point $\boldsymbol{p}^{\prime}_i$ in the reference view, if visible, must lie on the corresponding epipolar line $\mathbf{l}_i$.
Therefore, we opt to find the corresponding feature in the reference view along the epipolar line, which significantly reduces the search space and the memory required for subsequent computations.

We assume that the synthesized novel view images have the same camera intrinsic parameters $\mathbf{K}$ as the reference image, as being commonly set for the NVS task.
Specifically, given the relative camera rotation $\mathbf{R}$ and translation $\boldsymbol{t}$ from the reference image to the target image, for each point $\boldsymbol{p}_i$ in the target image, the corresponding epipolar line  $\mathbf{l}_i$ is:
\begin{equation}
\label{eq:epipolar}
    \boldsymbol{l}_i = \mathbf{R} [\boldsymbol{t}]_{\times} \mathbf{K}^{-1} \boldsymbol{p}_i,
\end{equation}
where $\mathbf{l}_i$ is the epipolar line of $\boldsymbol{p}_i$ in the reference image, and $[\boldsymbol{t}]_{\times}$ is the skew-symmetric matrix representation of $\boldsymbol{t}$.
Despite the unknown exact camera focal length $f$, the computation of the epipolar lines $\mathbf{l}_i$ remains feasible, as the computation can be independent of $f$ (see Supplementary Material for proof).

Subsequently, we sample a set of points denoted as $\boldsymbol{p}^{\prime} \in P^{\prime}$ along the epipolar line, specifically along the direction of the image width, at intervals of each feature pixel. Note that some sample points may be outside the feature plane and will be masked during the similarity calculation and feature retrieval process.

\subsection{Corresponding Point Searching}\label{sec:fine_locate}
\noindent \textbf{Paired Feature Acquiration.}
The epipolar sampling operation essentially reduces the search space of the corresponding points. However, how to more accurately locate the actual corresponding point in the epipolar line remains unsolved. 
Previous works~\cite{dift, dino-sd} show that the features extracted by diffusion models show good semantic correspondence between two input images. 
Thus, a plausible approach is to seek the corresponding position in the reference image for each pixel in the target view by assessing the similarity between their respective features.
However, previous feature matching methods~\cite{dift, dino-sd} require feeding two paired images into the diffusion model separately and extracting their features for matching, making them unsuitable for our scenario where the target image is pending generation.
To address this, we employ DDIM inversion, as detailed in Sec.~\ref{subsec:ddim_inversion}, to acquire noise from the reference image.
This noise is then utilized to concurrently reconstruct the reference image alongside the denoising process of the target image, which we used to obtain the paired features.
Specifically, we progressively denoise the DDIM inversed initial noise $\boldsymbol{x}^R$ of the reference image using DDIM sampling and set the relative camera pose transformation as $[0, 0, 0]$ so that the reference image can be recovered. Meanwhile, we use the same $\boldsymbol{x}^R$ as the initial noise to generate the target view.
We can then obtain paired features by retrieving the features in the same denoising step and at the corresponding layer of both the input and target generation branches.
Since the sampled point in the epipolar line is in the sub-pixel location, we employ bilinear interpolation to obtain the feature value of each point $\boldsymbol{p}^{\prime}_i$ in the epipolar line.
We then analyze the similarity of the corresponding intermediate feature of the target and reference branch.

\noindent\textbf{Computing Epipolar Attention.}
We now have access to the paired features of the input and target images. However, the specific features within the U-Net structure to utilize, as well as the methodology for calculating their similarity, remain unclear.
Previous feature matching methods~\cite{dift, dino-sd} calculate the similarity of output feature $\boldsymbol{F}$ of the attention block use cosine similarity $\operatorname{CosSim}(\boldsymbol{F}_{tgt}(\boldsymbol{q}), [\boldsymbol{F}_{ref}(\boldsymbol{q}^{\prime})])$ ($[\cdot]$ is the bilinear interpolation operation), followed by a softmax operation.
However, our experiments reveal that the similarity derived from these features does not align well with our intended application. The resulting similarity map exhibits a relatively uniform distribution, indicating insufficient localization of the desired corresponding location and inadequate corresponding feature aggregation (see Fig. C.1. in the supp. mat.).
It is important to note that the query and key features employed within the multi-head self-attention block are intended for similarity calculation, so we opt to use the query $\boldsymbol{Q}$ from the target branch and the key $\boldsymbol{K}$ from the reference branch to compute the similarity according to Eq.~\ref{eq:attn_mat}.
Such similarity scores can pinpoint the corresponding location (see Fig. C.1. in the supp. mat.).

\subsection{Reference Feature Injection}\label{sec:feature_use}
After finding the location of the corresponding point, we introduce how to use such information to constrain the generation process of the target image.
First, to neglect the necessity of further training or fine-tuning, we employ a simple parameter duplication strategy, as shown in Fig.~\ref{fig:arch}(c), in which \textit{we directly instantiated the epipolar attention block with the well-trained parameters of the self-attention block}.
Similar to the attention operation~\cite{transformer}, we use the weighted sum to aggerate the corresponding information in the reference image as follows:
\begin{equation}
    \hat{\boldsymbol{F}}_{src} = \sum_{p^{\prime} \in \mathcal{P}^{\prime}} \operatorname{sim}\left( \boldsymbol{Q}_{tgt}(p), \boldsymbol{K}_{ref}(p^{\prime})\right) \cdot \boldsymbol{F}_{\mathrm{src}}\left(p^{\prime}\right),
\end{equation}
where $\operatorname{sim}(\cdot)$ is the similarity calculation operation as Eq.~\ref{eq:attn_mat}.
Similar to the residual connection in the original U-Net block, we fuse the output feature from our epipolar attention block with the original self-attention block with a pre-defined weight parameter $\alpha$, which can be formulated as $\boldsymbol{F} = \alpha\hat{\boldsymbol{F}}_{src} + (1-\alpha)\hat{\boldsymbol{F}}$.

\begin{figure}[t]
    \centering
    \includegraphics[width=.7\linewidth]{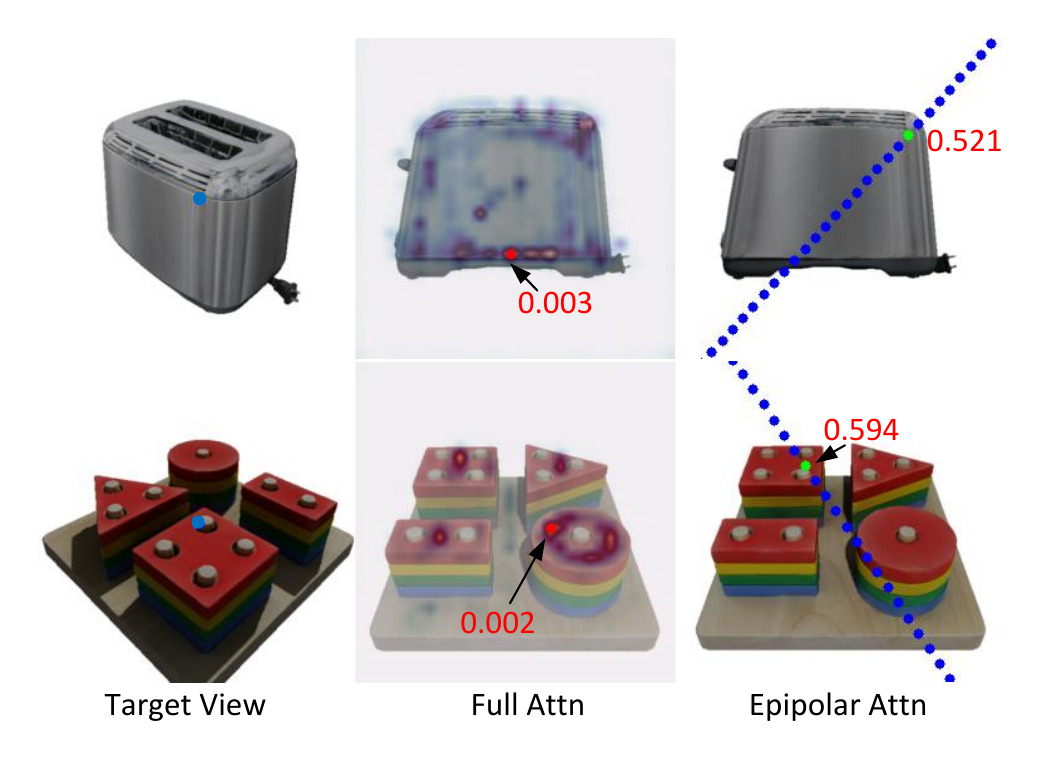}
    \vspace{-5mm}
    \caption{Comparison between our epipolar attention and the full attention. Our epipolar attention better locates and retrieves the corresponding information in the reference view.}
    \label{fig:full_attn_compare}
    \vspace{-6mm}
\end{figure}

\noindent\textbf{Attending Multi-Views at Once.}
While aggregating the overlapping feature from the input view improves the consistency between the output view and the input view, the consistency between different target views still is not well preserved as there are regions in the target images that are not visible to the input views. 
We further extend the epipolar attention to the multi-view setting to address this issue. 
Specifically, we generate multiple views $\Delta\boldsymbol{p}_i, i \in [1, N]$ in an auto-regressive manner. 
When synthesizing a specific novel view $\Delta\boldsymbol{p}_i$, we designate it as the target view. $M$ previous views, along with the input view, are considered context views, collectively containing specific information that overlaps with the target view. 
Subsequently, we apply epipolar attention to all context views and compute the average features derived from these views.
Fig.~\ref{fig:arch} (a) shows an example synthesis process for view $\Delta\boldsymbol{p}_i$.

\subsection{Discussion About the Epipolar Attention}\label{sec:attn_analysis}

\textbf{Comparison with Full Image Attention.}
An alternative to our epipolar attention mechanism is directly using full attention to gather corresponding information in the reference view, which finds the corresponding points in the full image. In contrast, our epipolar attention significantly reduces the search space for the corresponding point searching by introducing additional geometric priors. Illustrated in Fig.~\ref{fig:full_attn_compare}, our method exhibits sharper similarity scores and more precise localization of corresponding positions, resulting in a more effective retrieval of desired corresponding features. 
Thus, as shown in Tab.~\ref{tab:ablation}, epipolar attention performs better than full attention, especially when multiple reference views are employed.
Additionally, by reducing the search space, our epipolar attention significantly decreases memory consumption during the feature retrieval process. The space and time complexity of the epipolar attention is $O(L^3)$, while that of the full attention is $O(L^4)$, where $L$ is the length of the feature map.

\noindent\textbf{Comparison with Recent Methods.}
Some recent works, such as MVDream~\cite{mvdream}, SyncDreamer~\cite{SyncDreamer}, and Zero123++~\cite{zero123++}, also aim to improve the consistency of synthesized multi-view images. However, these methods require time-consuming re-training. Moreover, they constrain the camera pose during training, limiting their ability to synthesize images to a fixed set of camera poses. For example, MVDream~\cite{mvdream} can only synthesize images with four fixed camera views. In contrast, our method can synthesize consistent multi-view images with arbitrary camera poses without re-training.

Previous work, \ie, PGD~\cite{tseng2023consistent} also utilizes epipolar attention in the generation task. However, it differs from our method mainly in two aspects.
\textbf{1)} Our method aims to enhance baseline model consistency without tuning, while PGD treats epipolar attention as a network module requiring full network training, making it resource-intensive. 
These differences also lead to problem formulation in using epipolar constraints.
PGD computes per-pixel distances to the epipolar line as an additional weight map multiplied by the original attention matrix, thereby altering the original distribution of attention weights. 
Consequently, this approach is not suitable for a non-training pipeline. 
In contrast, we aim to \textit{locate} and \textit{retrieve} corresponding information from the reference views using the epipolar constraint to roughly approximate the correspondence, followed by sampling and soft fine-locating.
Thus, we avoid the need for time-consuming retraining. 
Furthermore, our method reduces GPU memory consumption compared to PGD, as PGD still utilizes full attention.
Inserting PGD's epipolar module into our pipeline yields inferior results and has no significant improvements over full attention (see Tab.\ref{tab:view32_free_compare} and Tab.~\ref{tab:ablation}).
\textbf{2)} To make the whole pipeline work without any fine-tuning, we invest considerable effort in its design, which is not explored in PGD. For instance, we provide insights into how to generate input view features based on pre-trained Zero123, determine appropriate features for similarity computation, and how to extend epipolar attention to multi-view setting.

%% file: sec/4_experiments.tex
\section{Experiments}
\label{sec:experiments}

\begin{figure*}[t]
\vspace{-6mm}
    \centering
    \includegraphics[width=0.8\linewidth]{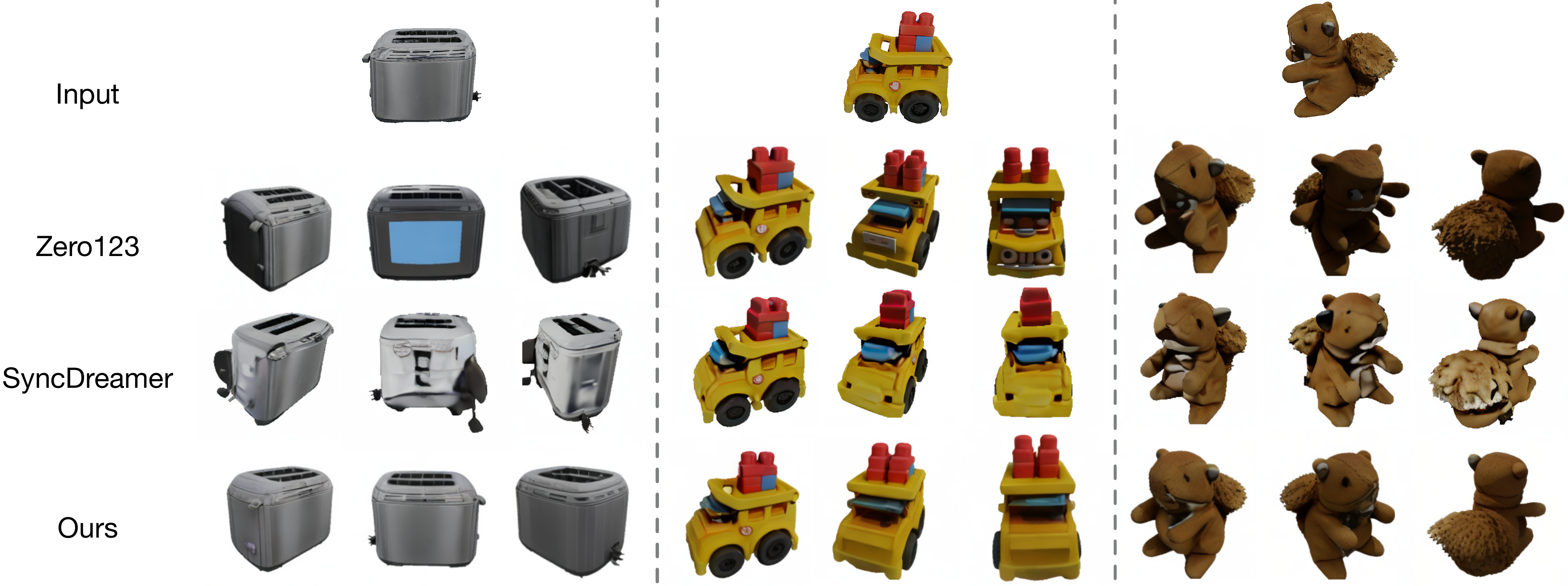}
    \vspace{-4mm}
    \caption{\textbf{Qualitative comparison} with the baseline for generating a sequence of novel view images.  
    The results demonstrate that our method synthesizes more consistent multi-view images compared to our baseline model (Zero123). In addition, compared to SyncDreamer, our method visually maintains better similarity to the conditioned image and appears more natural.}
    \label{fig:sota_compare}
\vspace{-5mm}
\end{figure*}

\subsection{Experimental Setups}
\textbf{Dataset.}
Following previous work~\cite{zero123, SyncDreamer}, we evaluate our work on the Google Scanned Object (GSO)~\cite{GSO} dataset to verify the zero-shot novel view image synthesis capability. 
We also provide results for additional datasets in the Supplementary Material.
Specifically, we randomly select 30 objects from the GSO dataset with various object categories. 
Unlike recent approaches~\cite{mvdream, SyncDreamer} that aim to enhance the consistency of novel view synthesis models by generating multiple fixed-view images, our method can generate images from any camera pose and any number of views. Therefore, we conduct experiments under different camera pose settings to validate our approach:
specifically, 
1) \textit{16-views with free camera pose}: for each object, we circularly render 16 views with the elevation angles ranging in $[-10\degree, 40\degree]$ and the azimuth angles are evenly distributed in $[0\degree, 360\degree]$. 
2) \textit{16-views with fixed camera pose}: We maintain a constant elevation angle of $30\degree$ and uniformly sample azimuth angles (same as SyncDreamer~\cite{SyncDreamer}).
3) \textit{32-views with free camera pose}: Similar to the first setting, but we sample 32 views.
It's important to note that our method does not require additional training or fine-tuning on any datasets.

\noindent\textbf{Metrics.}
To validate the effectiveness of our method, we mainly evaluate it based on three criteria:
1) \textit{Quality Score}. We evaluate the image quality of synthesized multi-view images by measuring their similarity with ground truth images. Following prior research~\cite{zero123, sparsefusion}, we report the similarity between the synthesized images and the ground truth images with standard metrics: PSNR, SSIM~\cite{ssim}, and LPIPS~\cite{lpips}.
2) \textit{Multi-view Consistency Score}. As the primary goal of our work is to improve the consistency of generated images, we also employ the 3D consistency score~\cite{3dim} to verify the consistency among the synthesized images. Specifically, we train an Instant-NGP~\cite{instant_ngp} with the input image and part of the synthesized novel view images of our model and evaluate the similarity between the remaining synthesized images and the rendered images of Instant-NGP. For the synthesized multi-view images of each object, we allocate $3/4$ for training and reserve the remaining $1/4$ for validation.
Intuitively, if the consistency of synthesized images is improved, the NeRF-like model will train a better object representation, and the re-rendered images will agree more with the validation images.
3) \textit{Input Consistency Score}. To assess the faithfulness of synthesized images in preserving the identity of the input condition image, we introduce the input consistency score. This score calculates the similarity of each synthesized image with the input condition image, utilizing the LPIPS metric.

In addition, we use synthesized multi-view images to train a neural 3D reconstruction model (NeuS~\cite{neus}) and report commonly used Chamfer Distances (CD) and Volume IoUs between the trained 3D model and the ground truth.

\noindent\textbf{Baselines.}
Given that our main goal is to improve the consistency of the trained baseline model without further fine-tuning, we mainly compare our approach with the used baseline model Zero123~\cite{zero123}. Additionally, we compare our method to the SOTA approaches such as PGD~\cite{tseng2023consistent} and SyncDreamer~\cite{SyncDreamer} using the same Zero123 base model.

\noindent\textbf{Implementation Details.}
We use the official checkpoint provided by Zero123~\cite{zero123}, which is trained on objaverse~\cite{objaverse} for 165,000 steps. We inject our epipolar attention layer after step $T=4$ and layer $L=10$ by default. We find that feature fusion weight $\alpha=0.5$, and the number of context views $M=2$ work better.

\input{tables/view16_free}
\input{tables/view16_fixed}

\subsection{Comparison With Baseline Models}
The quantitative comparison on three settings are shown in Tab.~\ref{tab:view16_free_compare}, Tab.~\ref{tab:view16_fxied_compare}, and Tab.~\ref{tab:view32_free_compare}. The qualitative comparison is shown in Fig.~\ref{fig:sota_compare}.

\input{tables/view32_free}
\input{tables/ablation}

\begin{figure*}[ht]
    \centering
    \begin{minipage}{0.65\textwidth}
        \centering
        \includegraphics[width=0.95\linewidth]{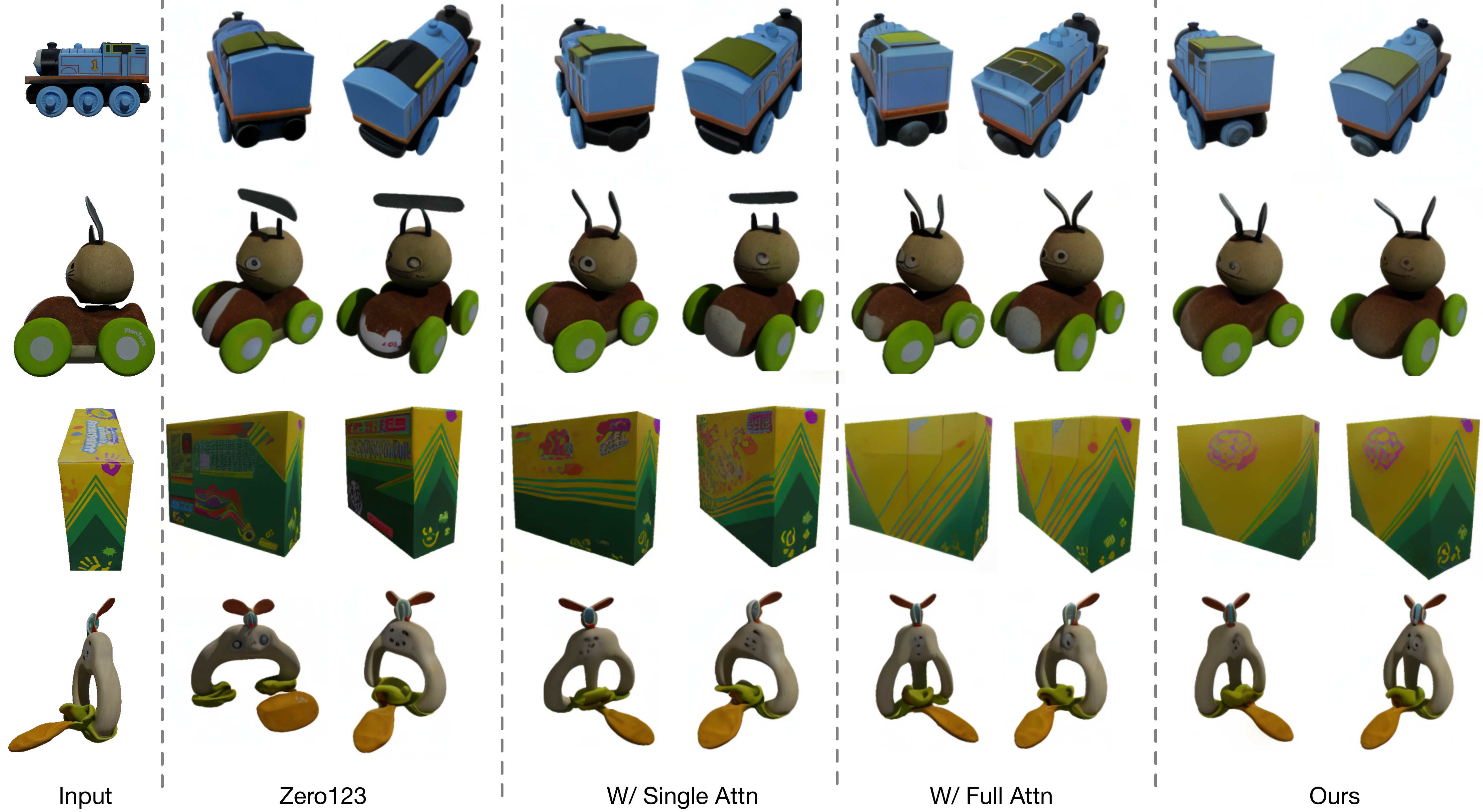}
        \vspace{-2mm}
        \captionof{figure}{Qualitative Comparison for different design choices. Our method, employing multi-view epipolar attention, demonstrates the best consistency.}
        \label{fig:ablation}
    \end{minipage}\hfill
    \begin{minipage}{0.33\textwidth}
        \centering
        \includegraphics[width=0.8\linewidth]{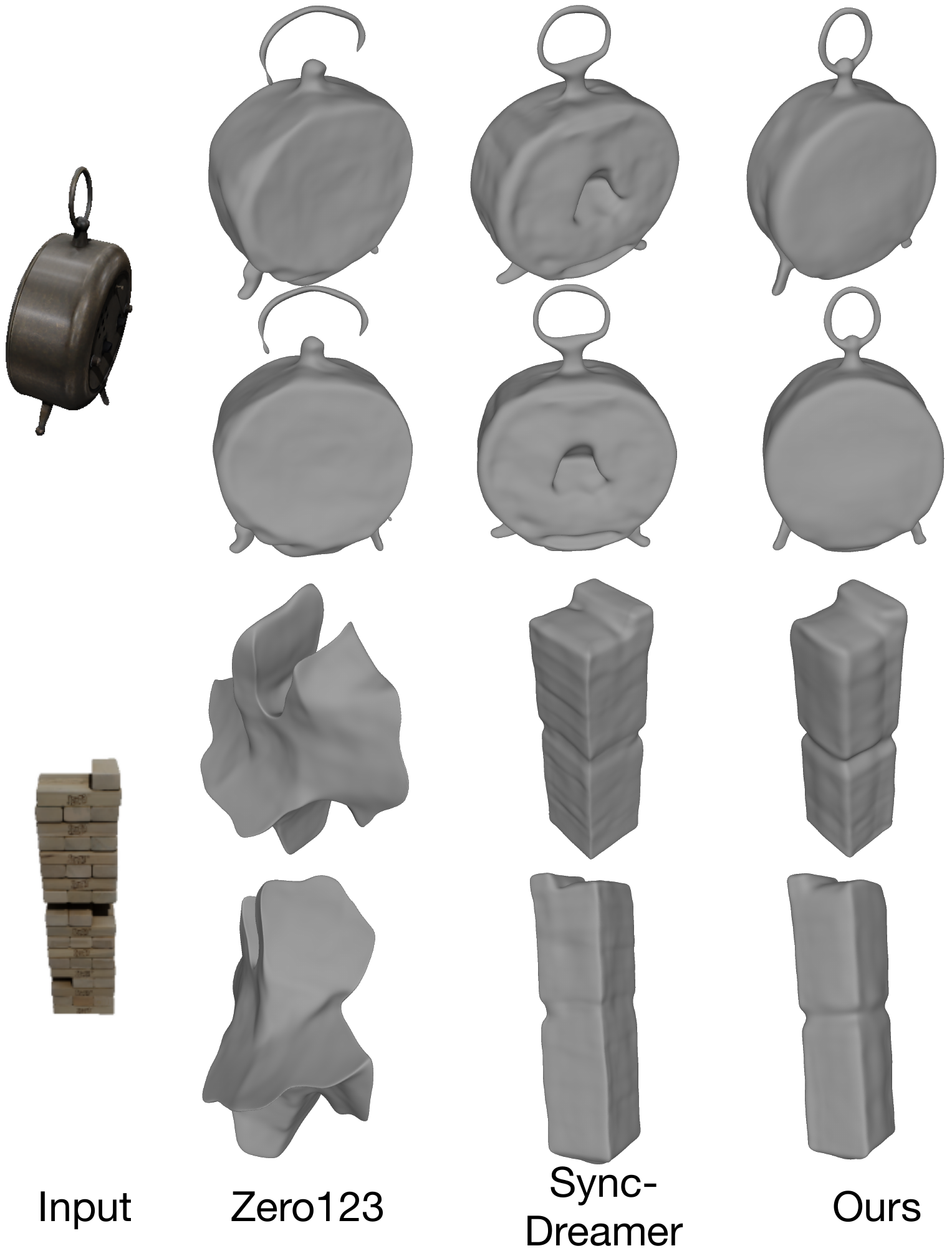}
        \vspace{-3mm}
        \caption{Our method shows better direct 3D reconstruction~\cite{neus}.}
        \label{fig:neus}
    \end{minipage}
    \vspace{-5mm}
\end{figure*}

\noindent\textbf{Multi-view Consistency.}
Tab.~\ref{tab:view16_fxied_compare} presents the 3D consistency scores compared to our baseline model (Zero123) and SyncDreamer. The results indicate a significant improvement across all three metrics achieved by our method when compared with Zero123.
While our method exhibits a marginally lower numerical consistency score compared to SyncDreamer, it enables the synthesis of images with arbitrary camera poses.	
This capability is illustrated in Tab.~\ref{tab:view16_free_compare}, where our method consistently enhances consistency with changes in camera pose settings, whereas SyncDreamer fails to do so and exhibits inferior results compared to Zero123.
Furthermore, our method facilitates the synthesis of multi-view images with any number of camera views. This versatility is demonstrated in Tab.~\ref{tab:view32_free_compare}, where our method continues to achieve significant improvements in consistency scores, while SyncDreamer is unable to operate under such conditions.	

Meanwhile, Fig.~\ref{fig:sota_compare} provides a qualitative comparison with the baseline. While both our method and SyncDreamer enhance consistency, our method visually preserves better similarity to the input image, including color and texture details. The input consistency score further corroborates this.

\noindent\textbf{Image Quality.}
While our primary goal centers around enhancing the consistency of synthesized multi-view images, we also evaluate the image quality by comparing the similarity with the ground truth images. The results shown in Tab.~\ref{tab:view16_free_compare}, Tab.~\ref{tab:view16_fxied_compare}, and Tab.~\ref{tab:view32_free_compare} indicate that our method also enhances the image quality under different settings besides improving the consistency.
Moreover, our method shows better image quality compared with SyncDreamer even in the 16-view setting with fixed camera pose.

\noindent\textbf{Input Consistency.}
Input consistency terms whether the results align with the input image.
Fig.~\ref{fig:sota_compare} illustrates that both our method and SyncDreamer enhance multi-view consistency. However, the color and texture details of SyncDreamer's results diverge from the input image and appear visually unnatural.
This discrepancy is evident in the input consistency score presented in Tab.~\ref{tab:view16_fxied_compare}, indicating lower similarity with the condition image in the SyncDreamer results.	

\subsection{Ablation Study}
The overall quantitative results are shown in Tab.~\ref{tab:ablation}, and the qualitative comparisons are shown in Fig.~\ref{fig:ablation}.

\noindent \textbf{Full Attention \vs Epipolar Attention.}
The results presented in Tab.\ref{tab:ablation} and Fig.\ref{fig:ablation} demonstrate that our epipolar attention mechanism can synthesize more consistent multi-view images compared with full attention. Furthermore, our epipolar attention achieves a greater performance improvement compared to full attention when using multiple reference images. This could be attributed to the fact that our epipolar attention more effectively localizes target information, as depicted in Fig.~\ref{fig:full_attn_compare}, thereby reducing noise from the reference images. In the multi-view setting, where multiple reference images are utilized, this noise reduction becomes particularly crucial.
Moreover, it is noteworthy that the epipolar attention mechanism consumes less GPU memory compared to our baseline, as discussed in Sec.~\ref{sec:attn_analysis}.

\noindent \textbf{Attending Single-View \vs Multi-View.}
Applying the epipolar attention significantly improves the consistency between the input and target views. However, the consistency between different views in the unobserved regions of the input view is not well preserved.
After implementing our epipolar attention in the multi-view setting, the consistency across the generated multi-view images is further improved. The last row in Tab.~\ref{tab:ablation} shows that after applying our multi-view epipolar attention, the consistency score is further improved compared with the single-view setting. Besides, the qualitative result in Fig.~\ref{fig:ablation} also shows better consistency among different target views.

\input{tables/neus}

\vspace{-2mm}
\subsection{Downstream Application}
\vspace{-2mm}
To demonstrate the effectiveness of our method, we also applied it to the downstream 3D reconstruction task. Specifically, we trained the NeuS model~\cite{neus} directly using images synthesized by our method, Zero123, and SyncDreamer, respectively.
The quantitative results in Tab.~\ref{tab:neus} show that the consistent multi-view images synthesized by our method can significantly improve the 3D reconstruction quality.
Additionally, our method exhibits similar performance to SyncDreamer which requires time-consuming re-training.
The qualitative results in Fig.~\ref{fig:neus} show that it is challenging to train the NeuS model directly due to the lack of consistency in the images generated by Zero123. In contrast, our method generates more consistent multi-view images and, therefore, better reconstructs the geometry and texture details.
We show improvements on other downstream applications such as image-to-3D in the Supplementary Material.

%% file: tables/view16_free.tex
\begin{table}[t]
\centering
\caption{Comparison of multi-view consistency, image quality, and input consistency of synthesized multi-view images at the 16-view setting with free camera pose.}
\label{tab:view16_free_compare}
\vspace{-2mm}
\scalebox{0.6}{
\begin{tabular}{c ccc ccc c}
\toprule
              & \multicolumn{3}{c}{Multi-view Consistency} & \multicolumn{3}{c}{Quality Score} & \multicolumn{1}{c}{Input Consis.} \\
              \cmidrule(lr){2-4} \cmidrule(lr){5-7} \cmidrule(lr){8-8}
              & PSNR$\uparrow$  & SSIM$\uparrow$ & LPIPS$\downarrow$ 
              & PSNR$\uparrow$  & SSIM$\uparrow$ & LPIPS$\downarrow$ 
              & LPIPS$\downarrow$ 
              \\ \midrule

Zero123
& 15.225        & 0.645       & 0.408
& 14.255        & 0.747       &	0.208
& 0.303         
\\
SyncDreamer
& 14.830        & 0.626       & 0.434
& 12.650        & 0.713       &	0.254
& 0.317         
\\
Ours 
& \best{18.300}	& \best{0.734}	& \best{0.355}
& \best{14.947}	& \best{0.763}	& \best{0.191}
& \best{0.282}
\\

\bottomrule
\end{tabular}
}
\end{table}

%% file: tables/view16_fixed.tex
\begin{table}[t]
\vspace{-1mm}
\centering
\caption{Comparison of multi-view consistency, image quality, and input consistency at the 16-view setting with fixed camera pose as SyncDreamer~\cite{SyncDreamer}.}
\label{tab:view16_fxied_compare}
\vspace{-3mm}
\scalebox{0.6}{
\begin{tabular}{c ccc ccc c}
\toprule
              & \multicolumn{3}{c}{Multi-view Consistency} & \multicolumn{3}{c}{Quality Score} & \multicolumn{1}{c}{Input Consis.} \\
              \cmidrule(lr){2-4} \cmidrule(lr){5-7} \cmidrule(lr){8-8}
              & PSNR$\uparrow$  & SSIM$\uparrow$ & LPIPS$\downarrow$ 
              & PSNR$\uparrow$  & SSIM$\uparrow$ & LPIPS$\downarrow$ 
              & LPIPS$\downarrow$ 
              \\ \midrule

Zero123
& 16.556        & 0.682       & 0.378
& 14.592        & 0.750       &	0.207
& 0.305         
\\
SyncDreamer
& \best{22.424}        & \best{0.812}       & \best{0.268}
& 15.269        & 0.749       &	0.196
& 0.300         
\\
Ours 
& 21.151	& 0.780	& 0.302
& \best{15.293}	& \best{0.764}	& \best{0.184}
& \best{0.287}
\\

\bottomrule
\end{tabular}
}
\vspace{-4mm}
\end{table}

%% file: tables/view32_free.tex
\begin{table}[t]
\centering
\caption{Comparison of multi-view consistency and image quality scores of synthesized multi-view images at the 32-view setting with free camera pose.}
\vspace{-3mm}
\label{tab:view32_free_compare}
\scalebox{0.7}{
\begin{tabular}{c ccc ccc}
\toprule
              & \multicolumn{3}{c}{Multi-view Consistency} & \multicolumn{3}{c}{Quality Score} \\
              \cmidrule(lr){2-4} \cmidrule(lr){5-7}
              & PSNR$\uparrow$  & SSIM$\uparrow$ & LPIPS$\downarrow$ 
              & PSNR$\uparrow$  & SSIM$\uparrow$ & LPIPS$\downarrow$ 
              \\ \midrule

Zero123
& 16.515        & 0.694       & 0.378
& 15.142        & 0.733       &	0.211
\\
PGD~\cite{tseng2023consistent}
& 18.481        & 0.720       & 0.343
& 15.281        & 0.739       &	0.205
\\
Ours 
& \best{20.655}	& \best{0.792}	& \best{0.305}
& \best{15.268}	& \best{0.742}	& \best{0.203}
\\

\bottomrule
\end{tabular}
}
\vspace{-3mm}
\end{table}

%% file: tables/ablation.tex
\begin{table}[t]
\centering
\caption{Ablation Study on consistency score and quality score. Each of the different design choices is added to the baseline model.}
\vspace{-3mm}
\label{tab:ablation}
\scalebox{0.75}{
\begin{tabular}{c ccc ccc}
\toprule
& PSNR$\uparrow$  & SSIM$\uparrow$ & LPIPS$\downarrow$ \\ \midrule

Baseline (Zero123)
& 16.515        & 0.694       & 0.378
\\

+ Full Attention (Single)
& 18.208	& 0.749	& 0.346
\\

+ Epipolar Attention (Single)
& 18.514	& 0.761	& 0.342
\\

+ Full Attention (Multi)
& 19.511	& 0.784	& 0.312
\\

+ Epipolar Attention (Multi)
& \best{20.655}	& \best{0.792}	& \best{0.305}
\\

\bottomrule
\end{tabular}
}
\vspace{-6mm}
\end{table}

%% file: tables/neus.tex
\begin{table}[t]
\centering
\vspace{-1mm}
\caption{Comparison of 3D reconstruction results. Our method significantly improves the reconstruction quality.}
\vspace{-3mm}
\label{tab:neus}
\scalebox{0.7}{
\begin{tabular}{c cc}
\toprule
              &  Chamfer Dist.$\downarrow$  & Volume IoU$\uparrow$
\\ \midrule

            Zero123         & 0.017         & 0.819    \\
            SyncDreamer     & \best{0.013}         & \best{0.847}    \\
            Ours            & 0.014	& 0.842 \\

\bottomrule
\end{tabular}
}
\vspace{-5mm}
\end{table}

%% file: sec/5_conclusion.tex
\vspace{-2mm}
\section{Conclusion}
\vspace{-2mm}
\label{sec:conclusion}

In this paper, we propose a method to improve the consistency of multi-view images synthesized by a pose-guided diffusion model without any training or fine-tuning.
Specifically, for each pixel in the target view, we use epipolar attention to locate and retrieve features at corresponding locations in the input view and insert them into the generation process of the target view to enhance consistency. We also extend epipolar attention to the multi-view setting by synthesizing multiple views and retrieving information from the input and other target views.
Experimental results show that our method can improve the consistency of the generated multi-view images and further benefit downstream applications such as 3D reconstruction.

\textbf{Acknowledgement.} Botao Ye is partially supported by the ETH AI Center.

%% file: sec/X_suppl.tex
\clearpage
\setcounter{page}{1}
\maketitlesupplementary
\appendix
\counterwithin{figure}{section}
\counterwithin{table}{section}

\section{Epipolar Line Calculation}
\label{sec:epipolar}

Here we provide detailed proof that the final epipolar line  $\boldsymbol{l}_i$ is independent of the unknown focal length $f$.

Given the rotation matrix $\mathbf{R}$ and translation vector $\boldsymbol{t}$ between the two cameras, and the camera intrinsic parameters $\mathbf{K} = \begin{bmatrix} f & 0 & a \\ 0 & f & b \\ 0 & 0 & 1 \end{bmatrix}$, the epipolar line $\mathbf{l}_i$ in the reference image corresponding to a point $\boldsymbol{p}_i$ in the target image can be calculated as:
\begin{equation}
    \boldsymbol{l}_i = \mathbf{E} \tilde{\boldsymbol{p}}_i = \mathbf{R} [\boldsymbol{t}]_{\times} \tilde{\boldsymbol{p}}_i,
\end{equation}
where $\mathbf{E}$ is the essential matrix, $[\boldsymbol{t}]_{\times}$ is the skew-symmetric matrix representation of the translation vector $\boldsymbol{t}$, and $\tilde{\boldsymbol{p}}_i = \mathbf{K}^{-1} \boldsymbol{p}_i$ is the point $\boldsymbol{p}_i$ in the normalized image coordinates.

Now, expressing $\tilde{\boldsymbol{p}}_i$ in terms of \(\boldsymbol{p}_i\) and \(\mathbf{K}\):

\begin{equation}
    \begin{aligned}
        & \tilde{\boldsymbol{p}}_i=\mathbf{K}^{-1} \boldsymbol{p}_i \\
        & =\left[\begin{array}{ccc}
        1 / f & 0 & -a / f \\
        0 & 1 / f & -b / f \\
        0 & 0 & 1
        \end{array}\right]\left[\begin{array}{l}
        x \\
        y \\
        1
        \end{array}\right] \\
        & =\left[\begin{array}{c}
        x / f-a / f \\
        y / f-b / f \\
        1
        \end{array}\right] \\
        & =\left[\begin{array}{c}
        (x-a) / f \\
        (y-b) / f \\
        1
        \end{array}\right].
    \end{aligned}
\end{equation}

Substituting this into the equation for $\boldsymbol{l}_i$:

\begin{equation}
    \boldsymbol{l}_i = \mathbf{R} [\boldsymbol{t}]_{\times} 
        \left[\begin{array}{c}
            (x-a) / f \\
            (y-b) / f \\
            1
        \end{array}\right].
\end{equation}

\noindent Here, the coordinates $(x - a)/f$ and $(y - b)/f$ are simply scaled versions of the original image coordinates $x$ and $y$, and this scaling does not affect the linearity of the equation. Therefore, the final expression for $\boldsymbol{l}_i$ does not explicitly depend on $f$.

\section{Property of the Epipolar Attention}
\begin{figure}[t]
\centering
  \begin{subfigure}[t]{.36\linewidth}
    \centering
    \includegraphics[width=0.7\linewidth]{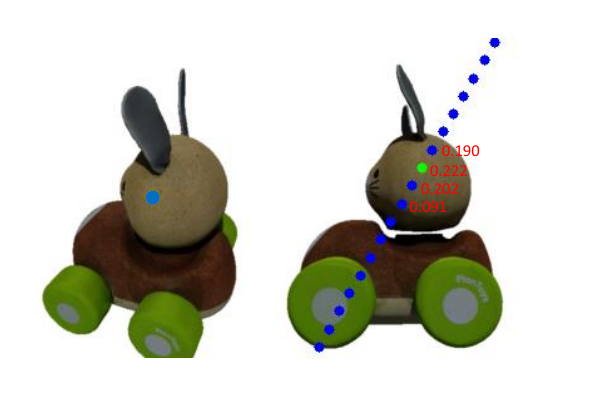}
    \caption{No clear correspondence}
  \end{subfigure}
  \hspace{1em}
  \begin{subfigure}[t]{.36\linewidth}
    \centering
    \includegraphics[width=0.7\linewidth]{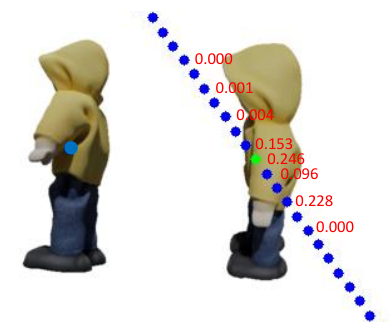}
    \caption{Occlusion}
  \end{subfigure}
  \hfill
\caption{When the occlusion occurs, or there is no clear geometric or semantic corresponding, epipolar attention tends to give multiple semantically similar points close similarity scores.}
\label{fig:attn_analysis}
\end{figure}

To better understand our epipolar attention mechanism, we performed a visual analysis of the attentional weights in various cases. In  Fig.~\ref{fig:attn_analysis}, two pairs of images show that our epipolar attention tends to give multiple semantically similar points close similarity scores when a point is occluded or when there is a lack of explicit geometric or semantic correspondence between the two points in the target and reference images. This behavior suggests that our method employs a broader range of contextual features, a favorable approach without explicit correspondences.

\section{Different Features for Similarity Calculation}
\begin{figure}[t]
    \centering
    \includegraphics[width=0.95\linewidth]{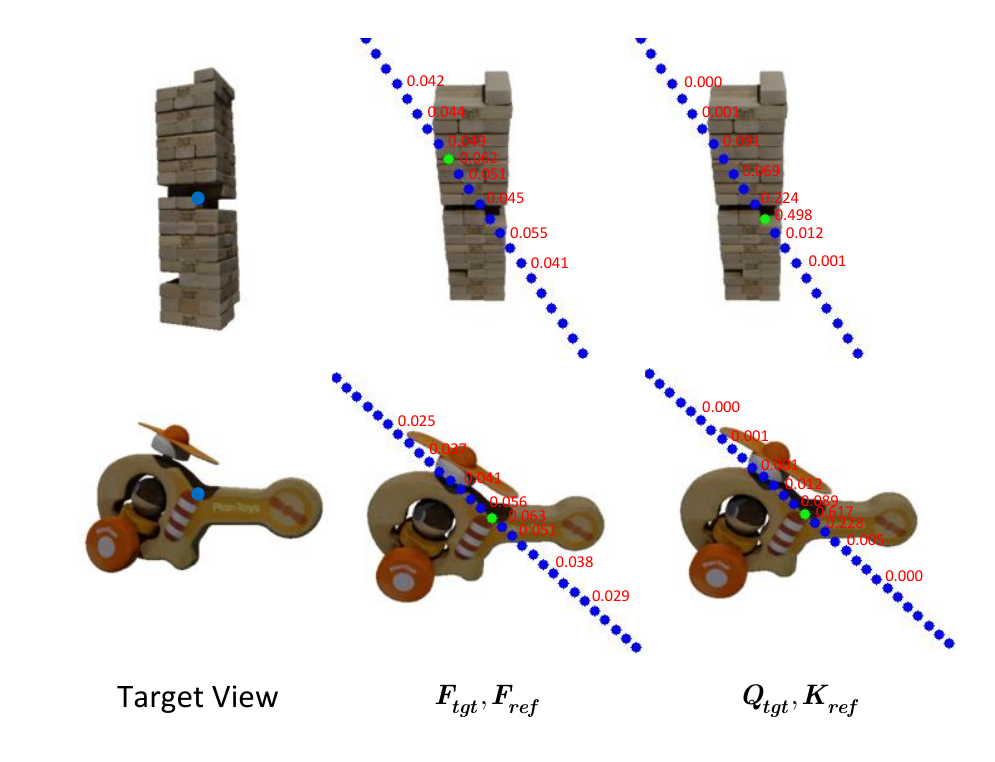}
    \caption{Similarity scores using different features. Similarity scores computed using queries and key features in the self-attention block are sharper and more accurate than those computed using the output features of the attention block.}
    \label{fig:attn_map}
\end{figure}

As discussed in Section 4.2 of our main paper, the similarity score derived from the output feature $\boldsymbol{F}$ of the attention block does not align well with our intended application, as it produces a relatively uniform similarity map. Instead, using the query $\boldsymbol{Q}$ from the target branch and the key $\boldsymbol{K}$ from the reference branch within the multi-head self-attention block provides a more accurate correspondence. This is illustrated in Figure~\ref{fig:attn_map}.

\section{Results on More Datasets}
\input{tables/objaverse}
We conduct experiments on the Objaverse dataset~\cite{objaverse}. Specifically, we randomly sample 100 objects from the Objaverse test set, utilizing the camera setting of 16-views with a fixed camera pose, which aligns with SyncDreamer's setup for fair comparison.
The results are presented in Tab.~\ref{tab:objaverse_view16_fix} and share the same conclusion with the exprimences on GSO~\cite{GSO} dataset.
Specifically, compared with our baseline model (Zero123), our method significantly improves the multi-view consistency, image quality, and input consistency on the Objaverse dataset. Compared with SyncDreamer, we achieve similar multi-view consistency but better image quality and input consistency. These results demonstrate the efficacy of our approach across different datasets.

\section{More Ablation Studies}
\input{tables/ab_num_views}
\input{tables/ab_feat_type}

\subsection{Number of Context Views}
The quantity of context views, denoted as $M$, may influence the consistency of synthesized multi-view images. Ablation studies are conducted to examine the impact of varying numbers of context views, and the results are presented in Tab.~\ref{tab:ab_num_views}. It is evident that in the absence of context views (our baseline), the consistency is poor. As the number of context views increases, the consistency improves. However, as the context number is continuously increased, the consistency score decreases. This decline may be due to significant relative camera pose transformations, resulting in smaller overlapping regions between two views. Retrieving information from these views may adversely affect performance.

\subsection{Effect of Using Different Features}
In Fig. 4 of our main paper, we visually compare the similarity scores obtained using different features, \ie, employing query key features within the self-attention blocks and output features of the self-attention layers. Here, a quantitative comparison is conducted to demonstrate the impact of employing distinct features. The results in Tab.~\ref{tab:ab_feat_type} illustrate that utilizing query key features shows better consistency performance than using the output features from the self-attention layers, as they better locate the corresponding features.

\input{tables/ab_overlap}
\subsection{Effectiveness on Different Overlap Ratios}
In Section 5 of our main paper, we present three different view sampling methods used in our experiments. These methods ensure that each view sufficiently overlaps with its neighboring views, facilitating the transmission of overlapping information.
Here, we vary the overlapping ratio between the target and input views during the single-view synthesis process to examine the impact of different overlapping ratios.
The results in Tab.~\ref{tab:ab_overlap} show that our method consistently demonstrates improvements over the baseline across various overlap ratios. Notably, even in scenarios where there is no overlap between the reference and target views, our method obtains performance gains over the baseline. This can be attributed to our approach of utilizing the DDIM inverted noise from the reference view as the initial noise for the target view, thereby incorporating additional information from the reference view.

\subsection{Other Hyperparameters}
In regards to the feature fusion weight $\alpha$, the step $T$, and the U-Net layer $L$ after which we inject our epipolar attention layer, we conduct preliminary tests with various values on a few numbers of objects, ultimately selecting those that yield more visually appealing results. We do not attempt to determine the optimal values across the entire test set, as this approach is impractical. Furthermore, it is acknowledged that different objects may necessitate distinct hyperparameter values for better performance.

\section{Application in Image-to-3D Task}
To further validate the effectiveness of our method on downstream applications, we apply our method to the image-to-3D task and compare the results with our baseline Zero123. Specifically, given a single image, we use the output noise of our method and Zero123 to distill the NeRF~\cite{nerf} training process. We follow the method proposed in DreamFusion~\cite{dreamfusion}; please refer to this paper for more details. The results in Fig.~\ref{fig:dreamfusion} show that our method generates 3D objects with better geometric and texture details, especially the parts that are not visible in the input view.

\section{Limitations}
\begin{figure*}[t]
    \centering
    \includegraphics[width=0.8\linewidth]{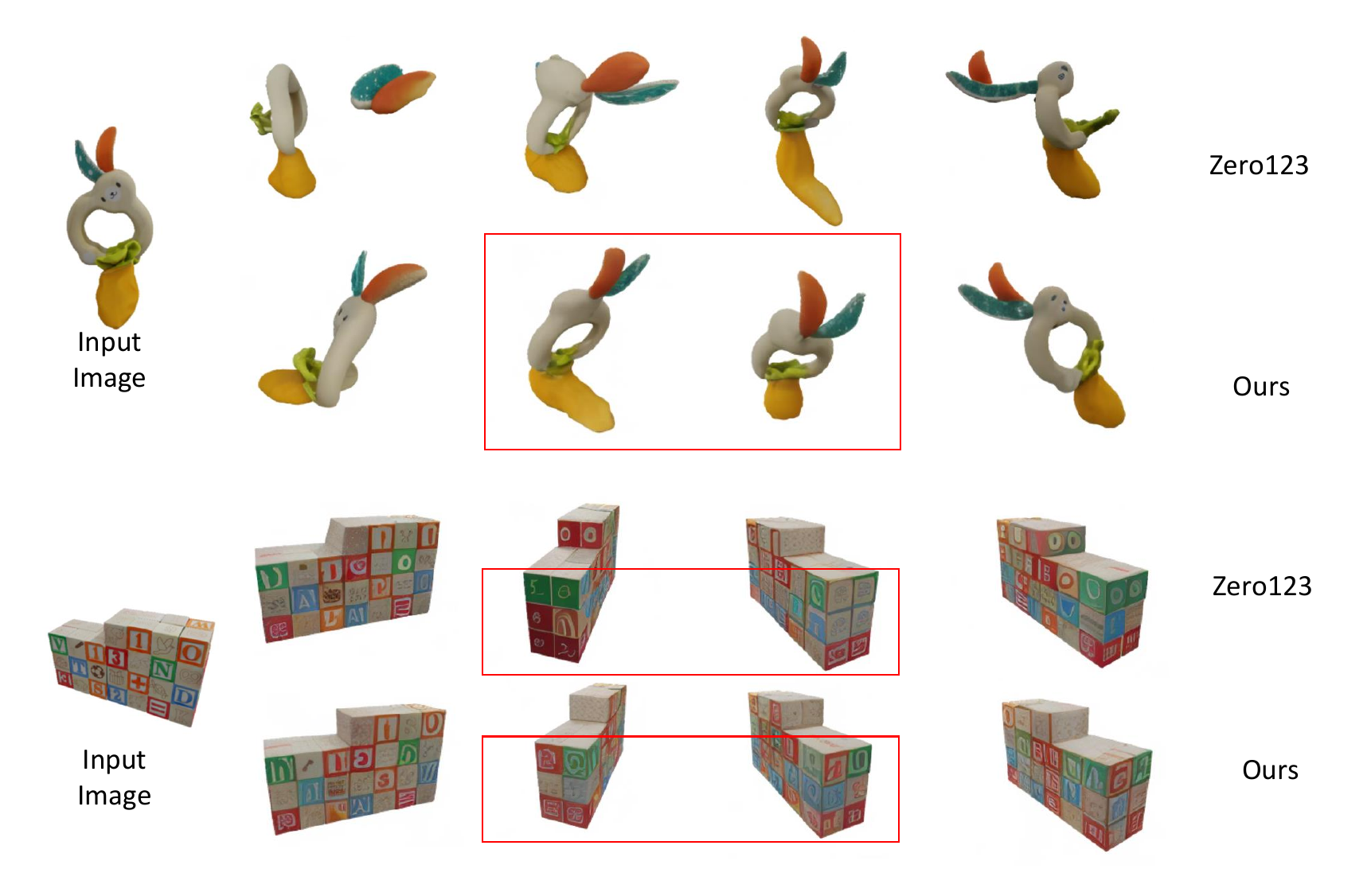}
    \caption{Failure cases. We provide an in-depth analysis of failure cases arising when the baseline model exhibits severe inconsistencies or when dealing with objects with complex textures.}
    \label{fig:failure}
\end{figure*}
Utilizing our epipolar attention to locate and retrieve corresponding information in the reference views enhances the consistency between generated multi-view images compared to the baseline model. Nevertheless, our method cannot ensure absolute consistency in the generated images due to the inherent probabilistic nature of the diffusion model, which remains unchanged. Employing multiple model runs and selecting superior results may further enhance consistency.

Here we further discuss failure cases in more detail.
1) Illustrated in the first set of images in Fig.~\ref{fig:failure}, our method encounters situations where severe inconsistencies exist in the baseline model, impeding its ability to well rectify these inconsistencies even when reference information is injected during the image generation process. In real-world applications, tuning the feature fusing weight $\alpha$ for a specific object may acquire better consistency results.
2) Illustrated in the second set of images in Fig.~\ref{fig:failure}, despite the substantial improvement in consistency achieved by our method in the generated multi-view images, our approach may encounter challenges maintaining absolute consistency, particularly when dealing with objects exhibiting complex textures. This limitation could stem from the inadequacy of the baseline model. Notably, our experiments demonstrate that even when a zero camera translation is provided to the model, it struggles to accurately reconstruct the input image in the presence of complex textures.

Besides, our auto-regressive generation pipeline naturally increases inference time. On a single NVIDIA A100, Zero123 generates a single image in 3 seconds, while our method takes 5 seconds. For 16 views, Zero123 takes 14 seconds due to batch processing, whereas our auto-regressive generation takes 55 seconds. However, considering the alternative of unaffordable re-training whenever a stronger baseline model becomes available, the runtime increase of our method is acceptable, as it significantly improves consistency and enables the generation of arbitrary views.

\section{More Visualization Results}

\begin{figure*}[t]
    \centering
    \includegraphics[width=0.8\linewidth]{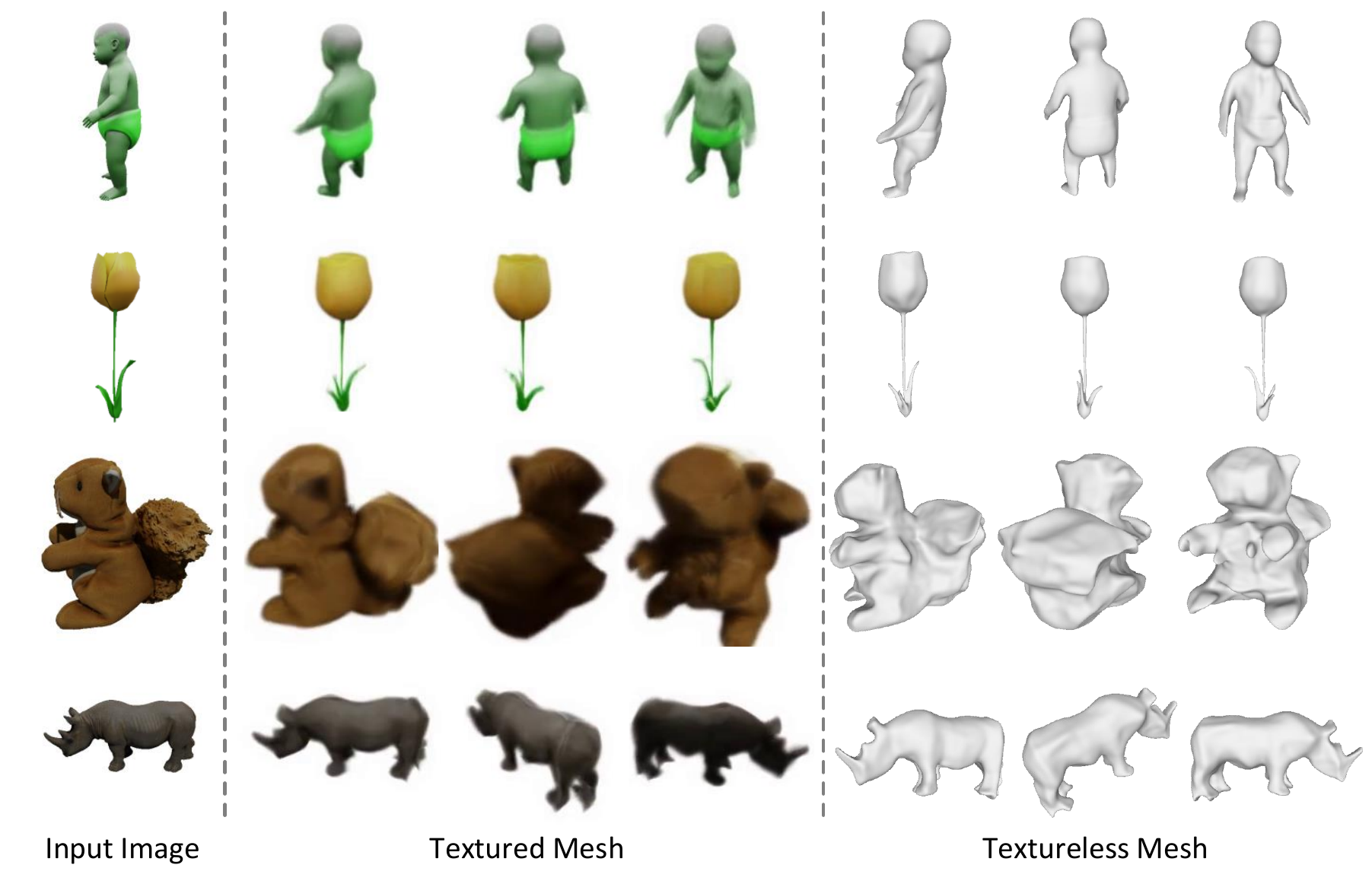}
    \caption{More 3D reconstruction results.}
    \label{fig:more_recon}
\end{figure*}

\begin{figure*}[t]
    \centering
    \includegraphics[width=0.95\linewidth]{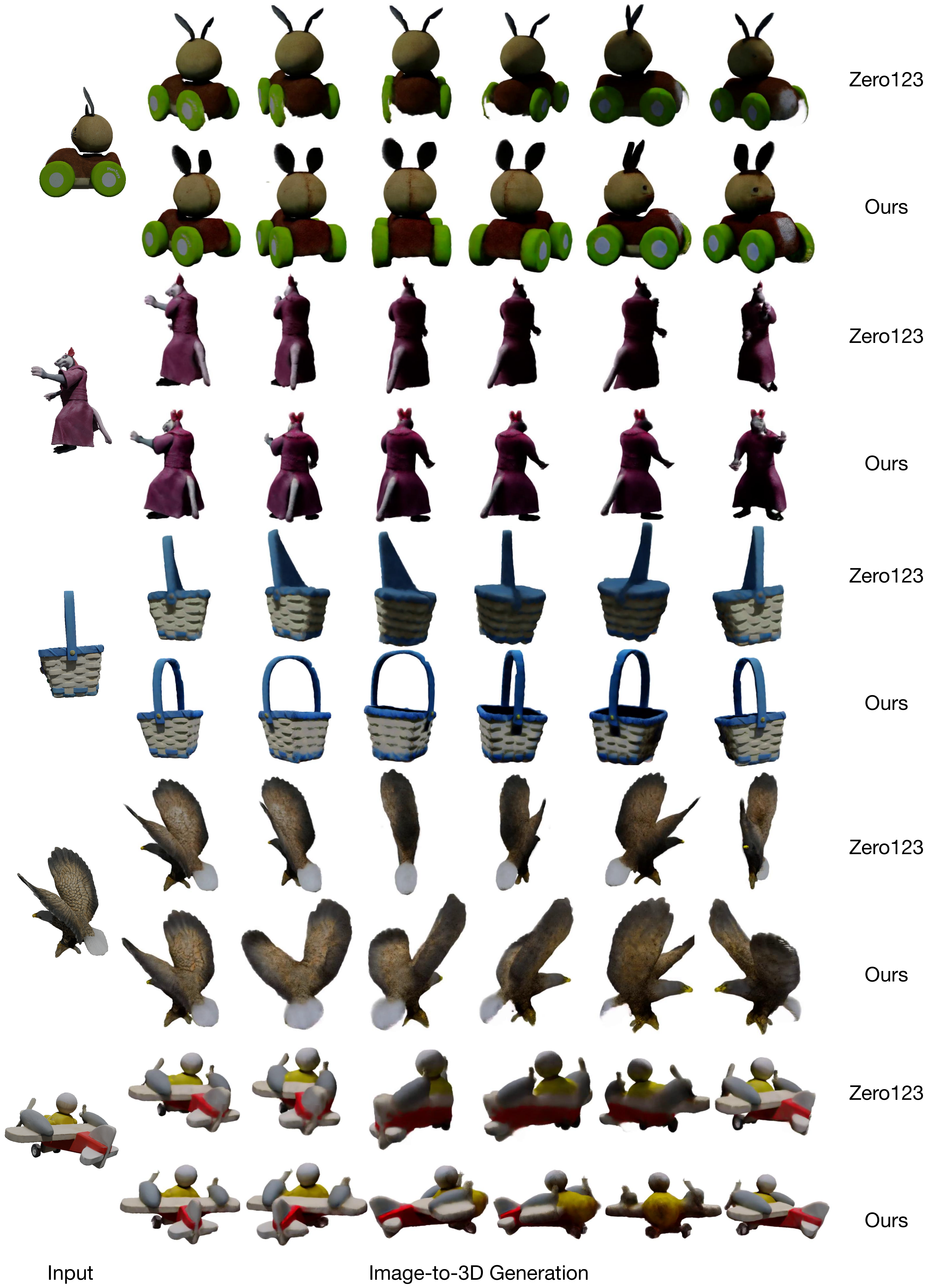}
    \caption{Image-to-3D generation results. In each group of images, the images in the first row depict results generated by the baseline model (Zero123), while those in the second row display results obtained from our approach. The results show that our method generates better 3D objects, especially the parts of the object not seen in the input view.}
    \label{fig:dreamfusion}
\end{figure*}

\noindent\textbf{More Reconstruction Results.}
We present additional 3D reconstruction results in Fig.~\ref{fig:more_recon}. These results illustrate that by increasing the consistency in the generated multi-view images, directly training 3D models using these images yields plausible 3D mesh representations.

\begin{figure*}[t]
    \centering
    \includegraphics[width=0.97\linewidth]{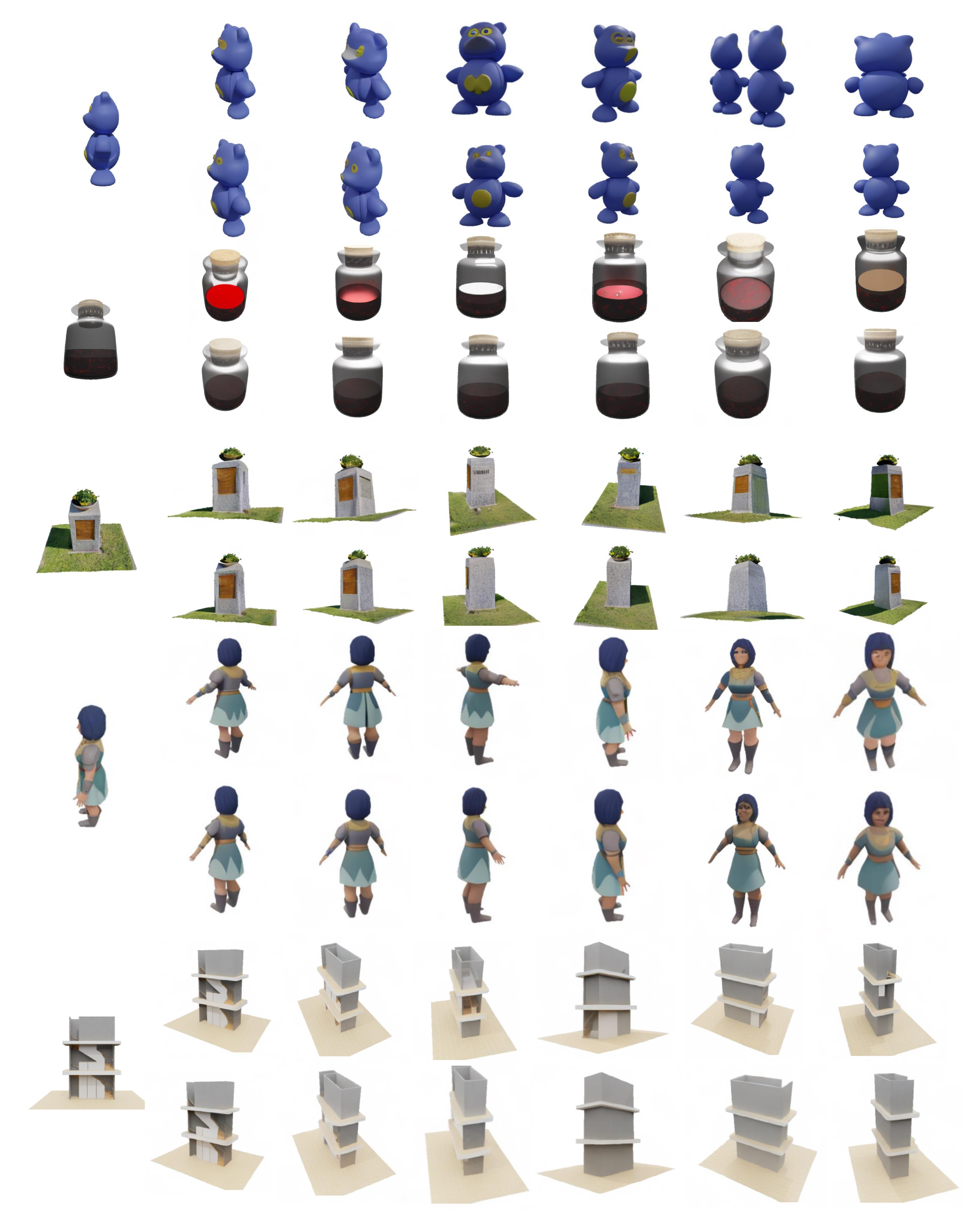}
    \caption{Qualitative comparison with the baseline for generating a sequence of novel view images on the Objaverse dataset.
    The images positioned on the left-hand side represent the input image. In each group of images, the images in the first row depict results generated by the baseline model (Zero123), while those in the second row display results obtained from our approach. 
    The comparison demonstrates that our method can generate multi-view images with higher consistency.}
    \label{fig:objaverse}
\end{figure*}

\begin{figure*}[t]
    \centering
    \includegraphics[width=0.92\linewidth]{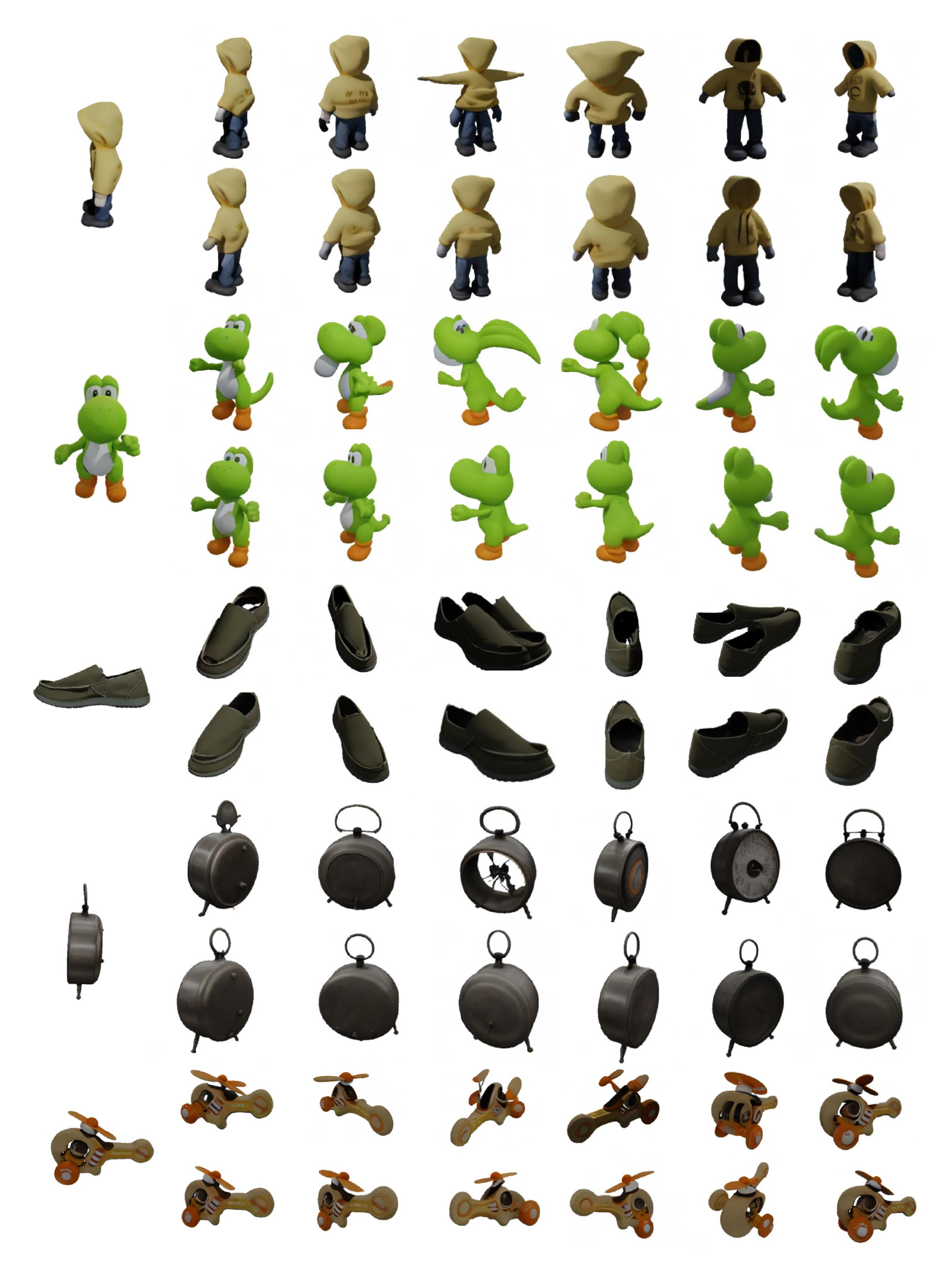}
    \caption{More Qualitative comparison with the baseline for generating a sequence of novel view images on the GSO dataset.
    The image placement aligns with Fig.~\ref{fig:objaverse}.}
    \label{fig:gso_more}
\end{figure*}

\noindent\textbf{More Qualitative Comparisons of Synthesized Multi-View Images.}
The results in Fig.~\ref{fig:objaverse} and Fig.~\ref{fig:gso_more} further provide comparisons of the multi-view images synthesized by the baseline model and our method.
In these two figures, the images positioned on the left-hand side represent the input image. In each group of images, the images in the first row depict results generated by the baseline model (Zero123), while those in the second row display results obtained from our approach.
The comparisons show that our method improves the consistency of generated multi-view images on different datasets.

\begin{figure*}[t]
    \centering
    \includegraphics[width=0.95\linewidth]{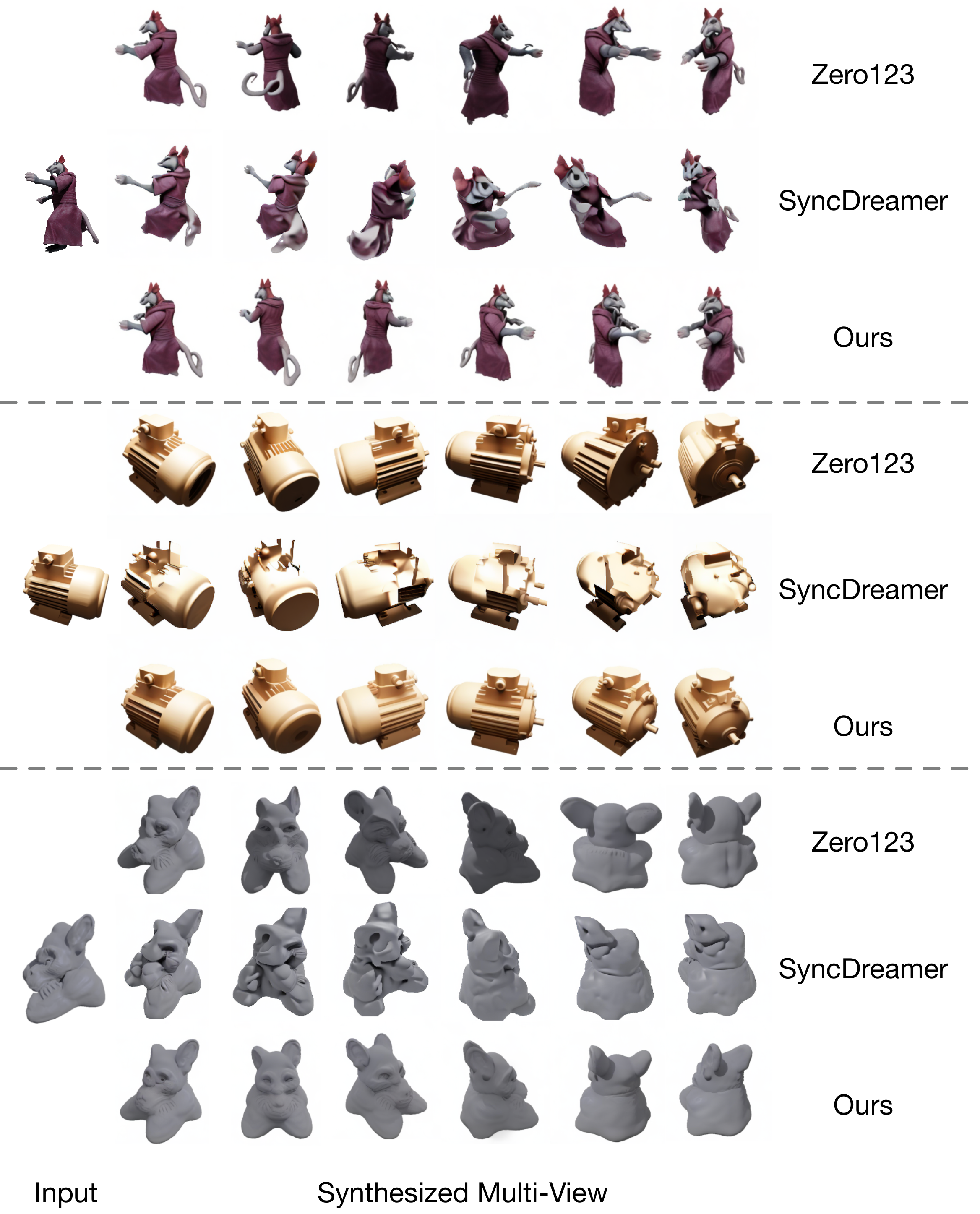}
    \caption{More Qualitative comparison with Zero123 and SyncDreamer. The results show that our method significantly improves multi-view consistency compared to Zero123, while also exhibiting better image quality compared to SyncDreamer.}
    \label{fig:more_compare}
\end{figure*}

The results in Fig.~\ref{fig:more_compare} provide additional comparisons between Zero123, SyncDreamer, and our method, demonstrating that our method significantly improves multi-view consistency compared to Zero123, while also exhibiting better image quality compared to SyncDreamer.

%% file: tables/objaverse.tex
\begin{table*}[t]
\centering
\caption{Comparison of multi-view consistency, image quality, and input consistency on Objaverse test set. The camera setting is the same as SyncDreamer~\cite{SyncDreamer}. The results show that our method has similar consistency scores to SyncDreamer, but higher quality scores and input consistency scores.}
\label{tab:objaverse_view16_fix}
\scalebox{0.95}{
\begin{tabular}{c ccc ccc c}
\toprule
              & \multicolumn{3}{c}{Multi-view Consistency} & \multicolumn{3}{c}{Quality Score} & \multicolumn{1}{c}{Input Consistency} \\
              \cmidrule(lr){2-4} \cmidrule(lr){5-7} \cmidrule(lr){8-8}
              & PSNR$\uparrow$  & SSIM$\uparrow$ & LPIPS$\downarrow$ 
              & PSNR$\uparrow$  & SSIM$\uparrow$ & LPIPS$\downarrow$ 
              & LPIPS$\downarrow$ 
              \\ \midrule

Zero123
& 19.271        & 0.769       & 0.324
& 19.533        & 0.808       &	0.162
& 0.265         
\\
SyncDreamer
& \best{23.827}        & \best{0.849}       & \best{0.257}
& 19.198        & 0.824       &	0.175
& 0.259         
\\
Ours 
& 23.341	& 0.830	& 0.263
& \best{21.147}	& \best{0.830}	& \best{0.144}
& \best{0.235}
\\

\bottomrule
\end{tabular}
}
\end{table*}

%% file: tables/ab_num_views.tex
\begin{table}[t]
\centering
\caption{Ablation study on the effect of the number of context views used. }
\scalebox{0.9}{
\begin{tabular}{c ccc}
\toprule
  & PSNR$\uparrow$     & SSIM$\uparrow$    & LPIPS$\downarrow$   \\
\midrule
0 (Baseline) & 16.556 & 0.682 & 0.378 \\
1 & 20.630 & 0.767 & 0.308 \\
2 (Ours) & \best{21.151} & \best{0.780} & \best{0.302} \\
3 & 20.937 & 0.772 & 0.311 \\
4 & 20.678 & 0.770 & 0.306 \\
5 & 20.450 & 0.773 & 0.305 \\
\bottomrule
\end{tabular}
}
\label{tab:ab_num_views}
\end{table}

%% file: tables/ab_feat_type.tex
\begin{table}[t]
\centering
\caption{Ablation study on using different features for matching. }
\scalebox{0.9}{
\begin{tabular}{c ccc}
\toprule
  & PSNR$\uparrow$     & SSIM$\uparrow$    & LPIPS$\downarrow$   \\
\midrule
Baseline        & 16.556 & 0.682 & 0.378 \\
Output Features & 20.045 & 0.771 & 0.327 \\
Query, Key      & \best{21.151} & \best{0.780} & \best{0.302}  \\
\bottomrule
\end{tabular}
}
\label{tab:ab_feat_type}
\end{table}

%% file: tables/ab_overlap.tex
\begin{table}[t]
\centering
\caption{The effectiveness of our method when the target view has different overlap ratios with the input view. Our method consistently demonstrates improvements over the baseline across various overlap ratios, even when no overlap exists.}
\scalebox{0.85}{
\begin{tabular}{ccccc}
\toprule
Overlap Ratio & 0.7    & 0.4    & 0.1    & 0(no overlap) \\
\midrule
baseline      & 17.089 & 15.296 & 14.354 & 13.350        \\
ours          & \best{17.214} & \best{15.678} & \best{14.603} & \best{13.448}  \\
\bottomrule
\end{tabular}
}
\label{tab:ab_overlap}
\end{table}